\documentclass[sigconf, natbib=true]{acmart}

\usepackage{mathtools} 
\usepackage{booktabs} 
\usepackage{tikz} 
\usepackage{xcolor}
\usepackage{cancel}
\usepackage{soul}
\usepackage{eucal}
\usepackage{url}
\usepackage{float}
\usepackage{amsmath,amsfonts}
\usepackage{amsthm, mathtools}
\usepackage{booktabs}
\usepackage{algorithm}
\usepackage{algorithmic}
\usepackage{multirow, multicol}
\usepackage{subfig}

\def\S{\mathcal{S}}
\def\A{\mathcal{A}}
\def\I{\mathcal{I}}
\def\P{\mathcal{P}_\mathcal{A}}

\def\s{\mathbf{s}}
\def\a{\mathbf{a}}

\DeclareMathOperator*{\argmax}{arg\,max}

\newtheorem{definition}{Definition}

\newtheorem{remark}{Remark}
\newtheorem{theorem}{Theorem}
\newtheorem{lemma}{Lemma}
\newtheorem{proposition}{Proposition}

\AtBeginDocument{%
  \providecommand\BibTeX{{%
    \normalfont B\kern-0.5em{\scshape i\kern-0.25em b}\kern-0.8em\TeX}}}

\copyrightyear{2021}
\acmYear{2021}
\setcopyright{acmcopyright}\acmConference[CIKM '21]{Proceedings of the 30th ACM International Conference on Information and Knowledge Management}{November 1--5, 2021}{Virtual Event, QLD, Australia}
\acmBooktitle{Proceedings of the 30th ACM International Conference on Information and Knowledge Management (CIKM '21), November 1--5, 2021, Virtual Event, QLD, Australia}
\acmPrice{15.00}
\acmDOI{10.1145/3459637.3482386}
\acmISBN{978-1-4503-8446-9/21/11}

\begin{document}
\fancyhead{}

\title{Revisiting State Augmentation methods for Reinforcement Learning with Stochastic Delays}


\author{Somjit Nath}
\affiliation{%
  \institution{TCS Research}
  \city{Mumbai}
  \country{India}}
\email{somjit.nath@tcs.com}

\author{Mayank Baranwal}
\affiliation{%
  \institution{TCS Research \\ IIT Bombay}
  \city{Mumbai}
  \country{India}}
\email{baranwal.mayank@tcs.com}

\author{Harshad Khadilkar}
\affiliation{%
  \institution{TCS Research \\ IIT Bombay}
  \city{Mumbai}
  \country{India}}
\email{harshad.khadilkar@tcs.com}


\begin{abstract}
  Several real-world scenarios, such as remote control and sensing, are comprised of action and observation delays. The presence of delays degrades the performance of reinforcement learning (RL) algorithms, often to such an extent that algorithms fail to learn anything substantial. This paper formally describes the notion of Markov Decision Processes (MDPs) with stochastic delays and shows that delayed MDPs can be transformed into equivalent standard MDPs (without delays) with significantly simplified cost structure. We employ this equivalence to derive a model-free Delay-Resolved RL framework and show that even a simple RL algorithm built upon this framework achieves near-optimal rewards in environments with stochastic delays in actions and observations. The delay-resolved deep Q-network (DRDQN) algorithm is bench-marked on a variety of environments comprising of multi-step and stochastic delays and results in better performance, both in terms of achieving near-optimal rewards and minimizing the computational overhead thereof, with respect to the currently established algorithms.
\end{abstract}




\maketitle

\section{Introduction}
Despite their enormous success, reinforcement learning (RL) in the basic Markov Decision Process (MDP) framework makes various restrictive assumptions that limit its applicability to real-world problems. Two of the most common simplifying assumptions that pose challenges in practical scenarios are absence of (a) \emph{observation delay}, i.e., a system's current state is always assumed to be available to the agent, and (b) \emph{action delay}, i.e., an agent's action has an immediate effect on the system's trajectory. It is well-known that the presence of delays in dynamical systems degrades the performance of the agents, resulting in undesirable behaviors of the underlying closed-loop systems and leading to instability~\cite{logemann1998destabilizing,patanarapeelert2006theoretical}.


Many real-world problems, including but not limited to congestion control~\cite{altman1999congestion}, control of robotic systems~\cite{imaida2004ground,jin2008robust}, distributed computing~\cite{hannah2018unbounded}, management of transportation networks~\cite{dollevoet2018delay}, and financial reporting~\cite{whittred1984timeliness} exhibit the undesirable effect of action and observation delays towards performance of the corresponding agents. Medical domain is another significant and probably the most pertinent domain in today's world where decision making is based on observing delayed state information. For instance, the decision to self-isolate and quarantine in case of suspected COVID-19 infection is based on the outcome of swab-based test, the result of which is significantly delayed (often by a day or two), since the labs tend to run large batches of swab tests together, not to mention the time taken by the labs to collect samples~\cite{larremore2020test,rong2020effect}. Similarly in biological systems, visual and proprioceptive information for motor control is significantly ($\approx$500ms) delayed during transmission through neural pathways~\cite{zelevinsky1998does}.

In recent years, deep reinforcement learning (DRL) has demonstrated enormous capability in learning complex dynamics and interactions in highly constrained environments~\cite{mnih2015human,gibney2016google,silver2016mastering,henderson2018deep}. Almost all the problems that DRL addresses are modeled using MDPs. By ignoring delay of agents, one can equivalently model the underlying problem as partially observable MDP (POMDP). POMPDs are generalizations of MDPs, however solving POMDPs without estimating hidden action states leads to arbitrary sub-optimal policies~\cite{singh1994learning}. Therefore, it becomes imperative to augment the delayed observation with the last few actions in order to ensure Markov property~\cite{altman1992closed,katsikopoulos2003markov,walsh2009learning,ramstedt2020reinforcement} in delayed settings. 

The reformulation allows an MDP with delays to be viewed as an equivalent MDP without delays comprising of the same state-transition structure as that of the original MDP, but at the expense of increased (augmented) state-space and relatively complex cost structure involving a conditional expectation. While the state-augmentation is essential to ensure Markov property, the resulting Bellman equation with conditional expectation in the equivalent cost structure requires vast computational resources for estimation. In particular, if the effect of an action or state observation is delayed by $d$ units at any stage, then the time-complexity of computation of the expected cost scales as $O(d|\mathcal{S}|^2)$~\cite{altman1992closed,katsikopoulos2003markov}, where $\mathcal{S}$ denotes the finite set of states. Moreover, the size of the ``augmented" state grows exponentially with $d$, i.e., $\mathcal{I}=\mathcal{S}\times\mathcal{A}^d$, and thus tabular learning approaches become computationally prohibitive even for moderately large delays. Here $\mathcal{I}$ and $\mathcal{A}$ represent the finite set of information states and actions, respectively. In view of these challenges, most of the existing work on designing reinforcement learning algorithms has largely worked with assumptions of constant delays and trying to handle delays in the environment either by adopting model-based methods (which are again computationally expensive and susceptible to sub-optimality) or by simulating future observations.

In this paper, we further extend the applicability of reinforcement learning algorithms to delayed environments on the following fronts: (a) We formally define and introduce the constant delay MDP (CDMDP) and stochastic delay MDP (SDMDP), and show that MDP with deterministic or stochastic delays can be converted into equivalent MDPs without delays. This is in contrast with semi-Markov Decision Processes (SMDPs), where the agent can wait for the current state to become observable or wait for the most recent action to get applied before taking subsequent actions; (b) The equivalence presents itself as a general framework for modeling MDP with delays as MDPs without delays within which the augmented states (information states) carry the necessary information for optimal action selection, the conditional expectation of the cost structure is significantly simplified; (c) A Delay-Resolved Deep Q-Network (DRDQN) reinforcement learning algorithm is developed which is shown to achieve near-optimal performance with minimal computational overhead. In particular, by using deep neural networks as nonlinear function approximators and disregarding a tabular framework, the computational complexity scales as $O(|\mathcal{S}|+d|\mathcal{A}|)$, i.e., it scales linearly with the delay $d$, and thus becomes amenable to work with.

\noindent\textbf{A word on notation}: A no-delay Markov Decision Process (MDP) is denoted by the tuple $\langle\mathcal{S},\mathcal{A},\mathcal{P}_\mathcal{A},r,\gamma\rangle$, where $\mathcal{S}$ and $\mathcal{A}$ represent the finite sets of states and actions, respectively. For each triple $\left(\mathbf{s},\mathbf{s}',\mathbf{a}\right)\in\mathcal{S}^2\times\mathcal{A}$, the probability that the system state transitions from $\mathbf{s}$ to $\mathbf{s}'$ under the effect of the action $\mathbf{a}$ is denoted by $p_\mathbf{a}\left(\mathbf{s}'|\mathbf{s}\right)\in\mathcal{P}_\mathcal{A}$. The immediate reward associated with each state-action pair $\left(\s,\a\right)$ is depicted by $r\left(\s,\a\right)$. The cumulative reward is discounted by a factor $0<\gamma<1$ and is given by the infinite sum $\sum_{t=0}^\infty\gamma^tr\left(\s_t,\a_t\right)$. Let $d$ represent the instantaneous delay in an agent's action or observation. We use $\I=\S\times\A^d$ to denote the set of augmented states. Note that in case of stochastic delay, $d$ is not constant and thus the definition of $\I$ changes accordingly; however, with a slight abuse of notation, $\I$ is used to represent the set of augmented states independent of the total delay $d$ with the understanding that its meaning is clear from the context. Finally, we use $d_\mathbf{o}$ and $d_\mathbf{a}$ to differentiate between the delays in observation and action, respectively, though, it is shown later that the two delays have similar effects on optimal decision making in delayed environments.

\section{Decision Making With Delays}\label{sec:DecMaking}
\begin{figure}
    \centering
    \includegraphics[width=0.42\textwidth]{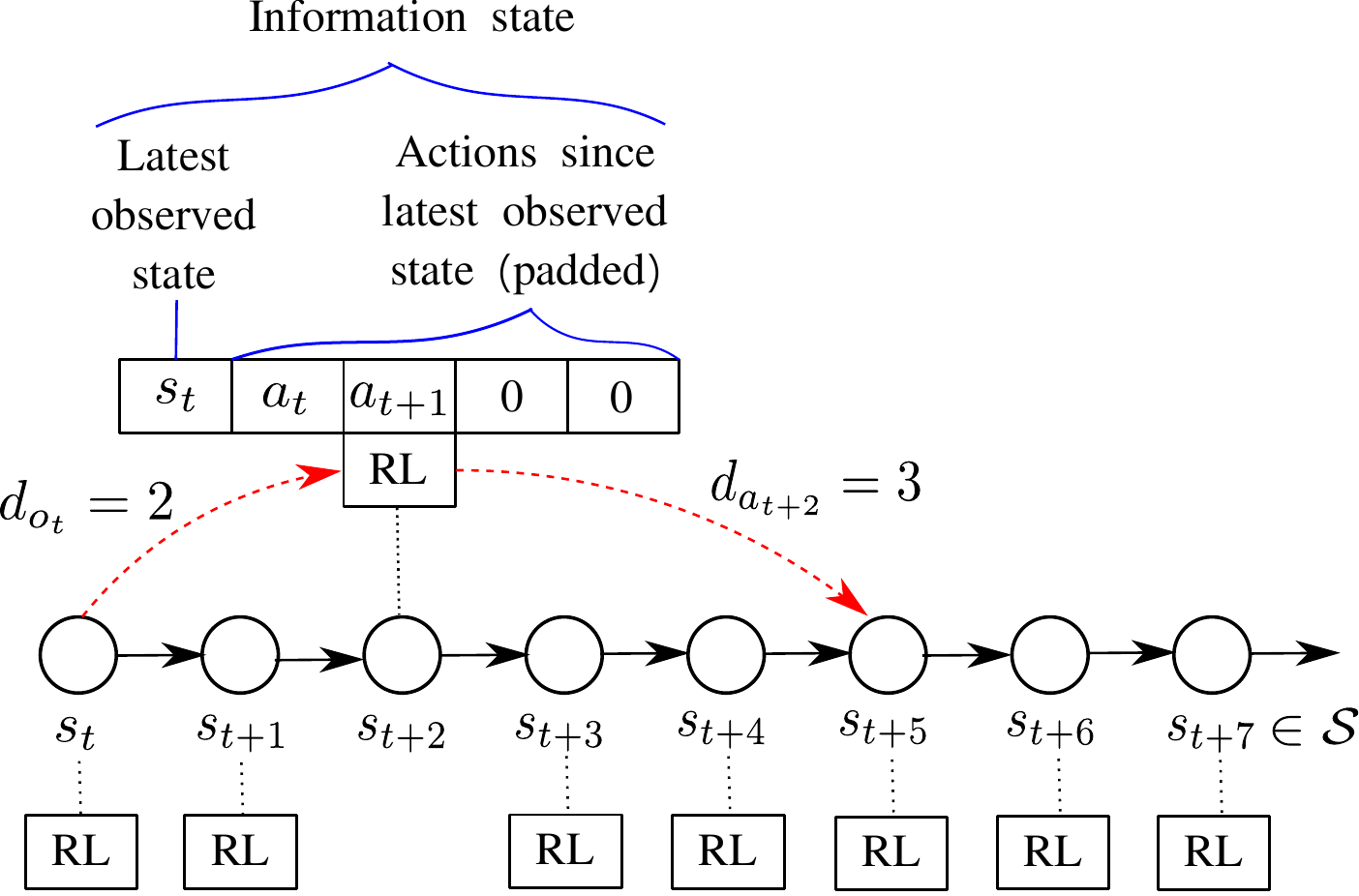}
    \caption{Graphical illustration of the problem with observation and actions delays. An action is computed in every time step, using the latest known state and a vector of all actions taken since that state was observed.}
    \label{fig:setup}
    \vspace{-1.5em}
\end{figure}
Fig.~\ref{fig:setup} is an illustration of an RL agent involved in sequential decision making in a delayed environment. At the beginning of each stage $t$, agent has access to the most recent observed state $\s_l$ with $l\leq t$, and the most recent sequence of actions $\left(\a_{l'},\a_{l'+1},\dots,\a_{t-2},\a_{t-1}\right)$ with $l'\coloneqq\min{\left(l,k\right)}$, where $\a_{k}$ is the most recent delayed action that got applied in an otherwise undelayed environment. It is easy to notice that in order to ensure Markov property in a stochastic delay setting, it is not only important to keep track of the most recent observed state, but also the time-instant at which it was observed first. Instead of naively waiting for the current state to become observable and the current action to get applied, agent must make informed decision at each step in order to minimize the total time taken to complete a given task.

\begin{figure}
    \centering
    \includegraphics[width=0.37\textwidth]{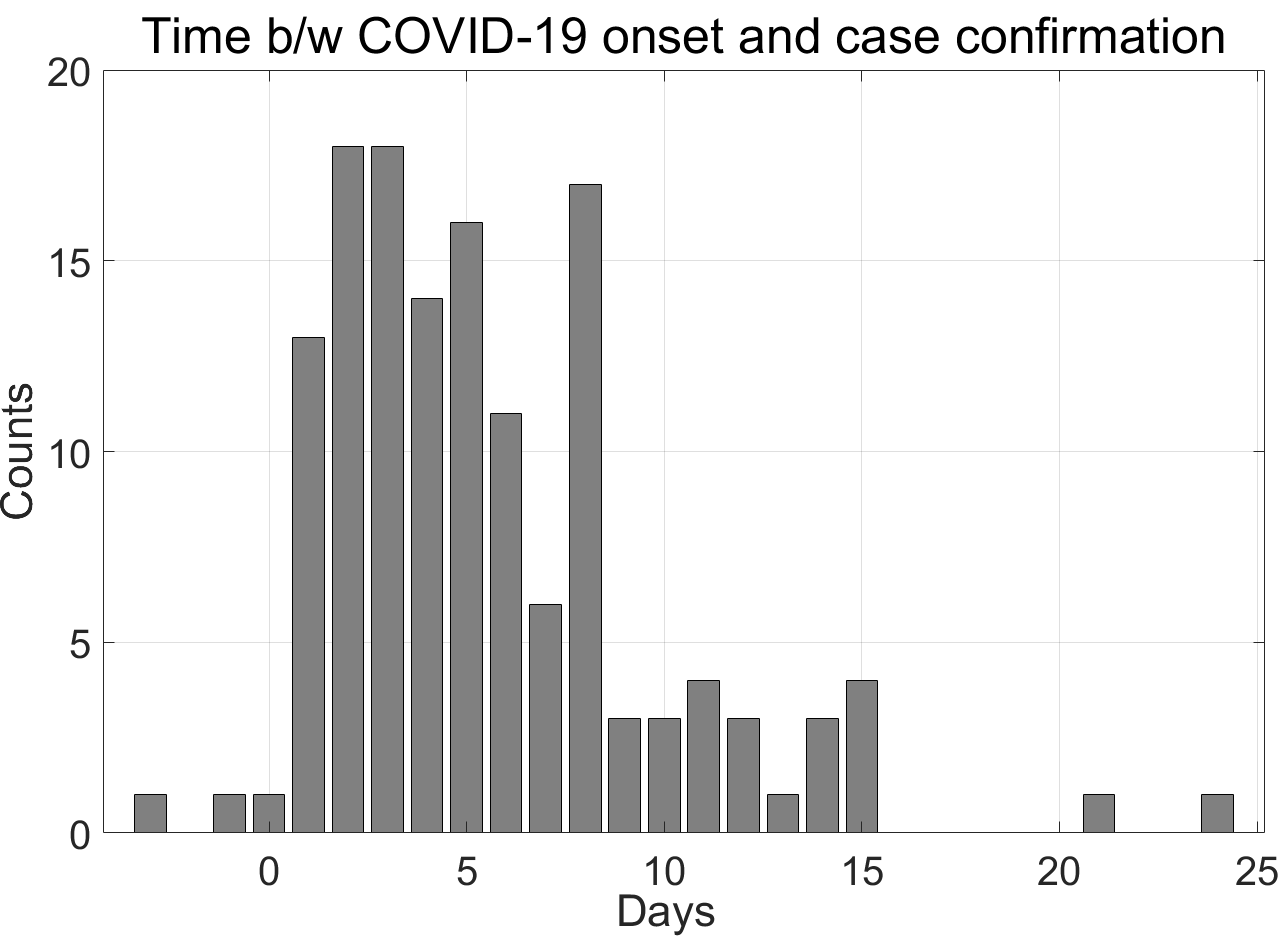}
    \caption{Histogram of the time between onset of COVID-19 symptom(s) and case confirmation for 139 patients from Basel-Landschaft, adopted from~\protect\cite{scire2020reproductive}.}
    \label{fig:delayCOVID}
    \vspace{-1.5em}
\end{figure}

One of the motivating examples of real-world delayed setting concerns with medical decision making for a suspected COVID-19 infection, where the mean time between symptom onset and a case being confirmed is estimated to be 5.6 days (standard deviation 4.2), see Fig.~\ref{fig:delayCOVID}. Therefore, based on a mean incubation period of 5.3 days~\cite{li2020early}, it may take up to 10.9 days on average for a case to be confirmed post-infection. Assuming that an agent starts developing some early flu-like symptoms, basing the decision making to self-isolate only after the availability of complete information (after 10.9 days) may result in further increase in infections around the subject. To overcome this concern, agent can instead resort to decision making in an MDP setting that takes into account the most recent observation (early symptoms, subject's proximity with potential confirmed cases, etc.) and action history (if the subject avoided close contacts with others during early symptomatic phase, agent's adherence to putting on a face-mask, etc.).  

Thus, the ability to perform decision-making in the presence of delays without having to wait for the effect of the current action to become observable is critical, particularly when the environment responds poorly to such `waiting' actions. Hence, it is extremely important to design algorithms with ability to perform real-time decision-making.

\section{Related Work}\label{sec:RelatedWork}
The main objective of this paper is to learn to perform optimal decision-making in delayed environments without having to wait for the observations to arrive or actions to get applied. It must be emphasized that the equivalent MDP formulation proposed in this paper with simplified reward structure and appropriately chosen information state need not have the same optimal reward as the original MDP, however, the two MDPs share the same optimal policy and we leverage this equivalence for optimal decision-making both in deterministic, as well as stochastic delayed environments. MDPs in delayed environments have been applied extensively in a variety of domains, particularly in optimal control of systems with delayed feedback~\cite{altman1992closed} and design of communication network in the presence of transmission delays~\cite{altman1999congestion}. Much of the prior work on delayed-MDPs~\cite{brooks1972markov,kim1985state,altman1992closed,kim1987partially,white1988note,bander1999markov} and POMDPs~\cite{sondik1978optimal} considered only constant observation delays, which was later extended to analyze constant delays in action and cost collection~\cite{bertsekas1995dynamic,altman1999congestion}. The notion of ``Constantly Delayed Markov Decision Process" was formally introduced in ~\cite{walsh2009learning}. MDPs with constant delays have largely been addressed using \emph{augmented} states in the literature~\cite{bertsekas1995dynamic,katsikopoulos2003markov}. However, as the size of the augmented state grows exponentially with the extent of delay, the augmented approaches find little applications due to intractability. A naive approach is to include memory-less methods that exploit prior information on the extent of constant delay~\cite{schuitema2010control} without the need for state augmentation. Consequently, model-based approaches were introduced to predict the current state in a delayed environment. These include involving estimation of transition probabilities in a multi-step delayed framework~\cite{10.1007/978-3-540-74958-5_41,chen2020delay}. However, model-based approaches are computationally expensive when compared with model-free approaches~\cite{otto2013curse}, let alone in the presence of delayed observations and actions. Fortunately, the recent advances in deep learning have facilitated practical implementation of augmented approaches as, unlike tabular methods, the complexity is only polynomial in the size of $\mathcal{A}$.~\cite{ramstedt2019real} recently formalized Real-Time Reinforcement Learning (RTRL), a deep-learning based framework that incorporates the effect of \emph{single-step} action delay. For small delays,~\cite{xiao2019thinking} analyzes the influence of incorporating the action-selection time towards correcting the effect of constant delay in an MDP.   ~\cite{ramstedt2020reinforcement} introduced partial trajectory sampling built on top of the Soft Actor Critic with augmented states in order to account for random delays in environments with significantly better performance. Most recently, \cite{dalal2021acting} also introduces use of forward models to improve performance for constant delays. 

The stochastic delayed environment in this work is fundamentally different from the stochastic setting described in~\cite{katsikopoulos2003markov}. \cite{katsikopoulos2003markov} makes impractical assumptions on asynchronous cost collection where observations arrive only in order, and cost is calculated only after the observation. Moreover, the amount of delay between any consecutive steps is non-negative, and the overall episodic delay can grow in no time. A more recent work on decision-making in stochastic delayed environments~\cite{ramstedt2020reinforcement} proposes a different solution for solving random delay problems by using augmented states along with the delay values for both action and observation. However, the authors in~\cite{ramstedt2020reinforcement} leverage the values of action and observation delays at each step to convert off-policy sampled trajectories into on-policy ones. Consequently, their approach is not suited for optimal decision-making in delay-agnostic environments discussed in this work. Detailed analysis on the relevant related work along with their merits and demerits are discussed further in Section~\ref{sec:comp}.

\section{Markov Decision Processes with Delays}\label{sec:delayMDPs}
We now formalize the general setting of an MDP with constant and stochastic delays in terms of the set of information states $\I$, action space $\A$, probability of state-transition $\P$, discount-factor $\gamma$, one-step reward $r\left(\cdot\right)$, delay in observation $d_{o}$, and delay in application of action $d_{a}$. It must be noted that from the point of view of the agent, action delay is functionally equivalent to observation delay (as shown later in Lemma~\ref{lem:equiv}). The effect of action (or observation) delay can better be understood through the following simple two-state MDP. The example below is adopted from~\cite{dalal2021acting}, however, \cite{dalal2021acting} considers a simple Markov chain where state-transitions are independent of the agent's actions. We suitably modify it to consider \emph{controlled} Markov chains or MDPs.
\begin{figure}
    \centering
    \includegraphics[width=0.32\textwidth]{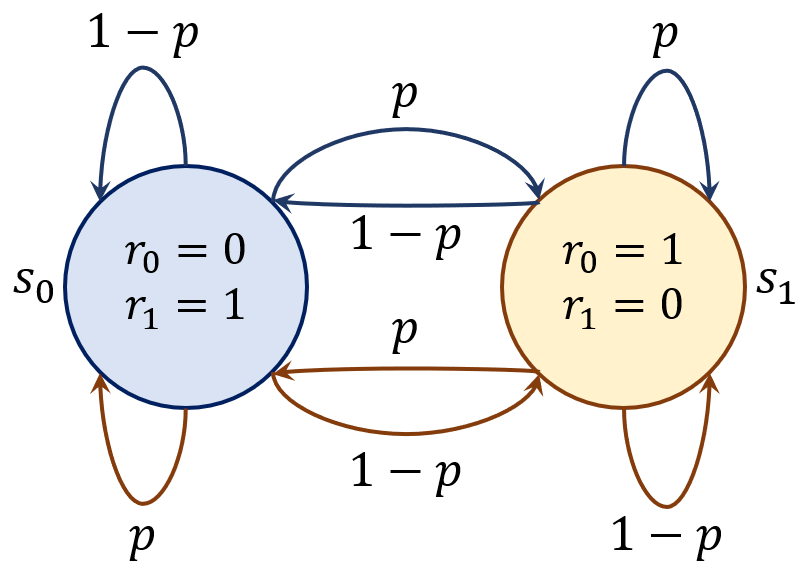}
    \caption{An illustrative example of how delay in action affects the optimal reward. Different colored arrows indicate state-transitions under different actions (Blue:0 and Brown:1).}
    \label{fig:2state}
    \vspace{-1.5em}
\end{figure}

\begin{proposition}\label{prop:2state}
    For any $p\in(.5,1]$, the optimal reward for the undelayed-MDP shown in Fig.~\ref{fig:2state} is greater than the optimal reward for the corresponding delayed version. Furthermore, if $p\to1$, i.e., in the limiting case of a deterministic MDP, the effect of delay becomes negligible. Thus, the example shown in Fig.~\ref{fig:2state} highlights the fundamental trade-off between stochasticity and delays.
\end{proposition}
\begin{proof}
    The proof uses the fact that the problem is symmetric in its states, and thus the optimal rewards from each state, as well as the corresponding optimal policies are also symmetric in the system states.
    We begin by making two key observations:\\
1. The problem is symmetric in the system states, i.e., the optimal reward $R^*(s_0)=R^*(s_1)$. Without loss of generality, we only consider computation of $R^*(s_0)$ in our analysis.\\
2. From symmetry, it can be further concluded that the optimal stationary policy $\pi_0^*(s_0) = \pi_1^*(s_1) \triangleq q$ and $\pi_0^*(s_1) = \pi_1^*(s_0) = 1-q$. It is later shown that the optimal policy necessitates $q=0$.

Let us first consider the case with no delay, then by definition:
\begin{align}\label{eq:Rtplus1}
    R_{t+1}^*(\s_t~&=s_0) = \sum\limits_{\s_{t+1}}r(\s_{t+1},0)\pi_0^*(s_0)p_0(\s_{t+1}|\s_t=s_0) \nonumber\\
    &\qquad \qquad + r(\s_{t+1},1)\pi_1^*(s_0)p_0(\s_{t+1}|\s_t=s_0) \nonumber\\
    R_{t+1}^*(s_0) &= r(s_0,0)\pi_0^*(s_0)(1-p) + r(s_1,0)\pi_0^*(s_0)p \nonumber\\
    &\quad + r(s_0,1)\pi_1^*(s_0)p + r(s_1,1)\pi_1^*(s_0)(1-p)\nonumber\\
    &= pq + p(1-q) = p,
\end{align}
since, $r(s_0,0)=r(s_1,1)=0$ and $r(s_0,1)=r(s_1,0)=1$. Furthermore, the transition probabilities are independent of time $t$, and thus $R^*_{t+1} = R^*_{t+k} = p\triangleq R^*$ for all $k\in\mathbb{Z}$.

Similarly, let us consider the optimal reward, $\tilde{R}_{t+2}^*$, under one-step delay, given by:
\begin{align}\label{eq:2step-1}
    \tilde{R}_{t+2}^*(\s_t~&\!=\!s_0) = \!\sum\limits_{\s_{t+2}}\!r(\s_{t+2},0)P(\s_{t+2}|\s_t\!=\!s_0,\a_t\!=\!0)\tilde{\pi}_0^*(s_0) \nonumber\\
    &\qquad + r(\s_{t+2},1)P(\s_{t+2}|\s_t\!=\!s_0,\a_t\!=\!1)\tilde{\pi}_1^*(s_0).
\end{align}
Note that from the problem description, $r(s_0,0)=r(s_1,1)=0$ and $r(s_0,1)=r(s_1,0)=1$, thus, \eqref{eq:2step-1} reduces to

\begin{align}\label{eq:2step-2}
    \tilde{R}_{t+2}^*(\s_t~&=s_0) = \tilde{q}P(\s_{t+2}\!=\!s_1|\s_t\!=\!s_0,\a_t\!=\!0) \nonumber\\
    &+ (1-\tilde{q})P(\s_{t+2}\!=\!s_0|\s_t\!=\!s_0,\a_t\!=\!1).
\end{align}
Again, since the transition probabilities are independent of $t$ and the optimal policies are stationary, $\tilde{R}_{t+2}^*(s_0) = \tilde{R}^*(s_0)$ for all $t$.

In order to evaluate the optimal reward in \eqref{eq:2step-2}, the corresponding two-step transition probabilities need to be computed. For brevity, the analysis below shows the computation of only the first term on the right-hand-side of \eqref{eq:2step-2}, (two-step transition probability under action 0). Fig.~\ref{fig:2statetran} shows the Markov chain for the aforementioned two-step transition along with corresponding one-step transition probabilities. Under action 0, the probability that the system state transitions from $s_0$ to $s_1$ is $p$ (and with probability $1-p$, the system remains in $s_0$). Likewise, the one-step transition probabilities are evaluated to reach the goal-state $s_1$.
\begin{figure}
    \centering
    \includegraphics[width=0.42\textwidth]{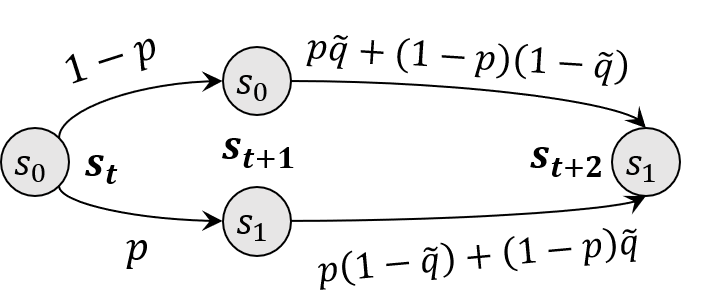}
    \caption{Illustration of 2-step transition probability from $s_0$ to $s_1$ under action 0.}
    \label{fig:2statetran}
    \vspace{-1.5em}
\end{figure}

From Fig.~\ref{fig:2statetran}, it follows that:
\begin{align}\label{eq:P1}
    P(\s_{t+2}\!=\!s_1|\s_t\!=\!s_0,0) &= (1\!-\!p)\left[p\tilde{q}+(1\!-\!p)(1\!-\!\tilde{q})\right] \nonumber\\
    &\quad +p\left[p(1\!-\!\tilde{q})+(1\!-\!p)\tilde{q}\right].
\end{align}
The transition probability from $s_0$ to itself under action 1 can similarly be obtained, and it turns out to be the same as in \eqref{eq:P1} for the given problem parameters. Thus, from \eqref{eq:2step-2} and \eqref{eq:P1}, the optimal return $\tilde{R}^*(s_0)$ is obtained as:
\begin{align*}
    \tilde{R}^*(s_0) &= (1\!-\!p)\!\left[p\tilde{q}+(1\!-\!p)\!(1\!-\!\tilde{q})\right] + p\!\left[p\!(1\!-\!\tilde{q})+(1\!-\!p)\!\tilde{q}\right] \\
    &= (1-\tilde{q}) + 2p(1-p)(2\tilde{q}-1).
\end{align*}
The return is maximized for a suitable choice of optimal policy $\pi^*$, leading to
\begin{align}\label{eq:R-tilde}
    \tilde{R}^*(s_0) &= \max\limits_{\tilde{q}} \left[(1-\tilde{q}) + 2p(1-p)(2\tilde{q}-1)\right] \nonumber \\
    &\leq p = R^*(s_0) \qquad\qquad \forall p\in[0.5,1].
\end{align}
Thus, from \eqref{eq:R-tilde}, it can be concluded that for any $p\in[0.5,1]$, the delayed reward $\tilde{R}^*(s_0)$ is always smaller than or equal to the undelayed reward $R^*(s_0)$. If $p\to 1$, the undelayed reward approaches the delayed reward with equality at $p=1$, indicating the fundamental trade-off between stochasticity and delays. The above analysis can be easily extended to consider any amount of delay (and not necessarily unit step delay).

\end{proof}

\subsection{Constant Delay Markov Decision Process}\label{subsec:CDMDP}
\begin{definition}
    A Constant Delay Markov Decision Process (CDMDP), denoted by the tuple $\left\langle\mathcal{S},\mathcal{A},\mathcal{P}_\mathcal{A},r,\gamma,d_o,d_a\right\rangle$, augments an MDP with state-space $\mathcal{S}\times\mathcal{A}^{d_o+d_a}$. Note that a policy $\pi$ is defined as a mapping $\pi:\mathcal{S}\times\mathcal{A}^{d_o+d_a}\to\mathcal{A}$.
\end{definition}
A CDMDP can inherit delays due to delay in both observation, as well as action. Intuitively, the two delays are functionally equivalent from the point of view of the agent. For instance, consider a scenario with no action delay, however, the observations are delayed by $d$ units of time. For an agent trying to select an optimal action at any time $t$, the corresponding reward is available only at time $t+d$, i.e., the action at time $t$ gets applied effectively at time $t+d$, as also happens in the action delay case. We formalize this intuitive argument in the form of Lemma~\ref{lem:equiv}.

It is shown in~\cite{altman1992closed} that CDMDP $\left\langle\mathcal{S},\mathcal{A},\mathcal{P}_\mathcal{A},r,\gamma,d,0\right\rangle$ can be reduced to an equivalent MDP $\left\langle\mathcal{I},\mathcal{A},\mathcal{P}_\mathcal{A},r,\gamma\right\rangle$ with $\mathcal{I}\coloneqq\mathcal{S}\times\mathcal{A}^d$ and suitably modified reward structure at any time $t$, given by $\tilde{r}\left(\mathbf{I}_t,\a_t\right)\coloneqq \mathbb{E}\left[r\left(\s_t,\a_t\right)|\mathbf{I}_t\right]$ for $\mathbf{I}_t\in\mathcal{I},\s_t\in\mathcal{S},\a_t\in\mathcal{A}$. Thus, from \cite{altman1992closed}, the information necessary for optimal action selection $\a_t$ at any instant $t$ is contained in $\mathbf{I}_t\coloneqq \left(\s_{t-d},\a_{t-d},\dots,\a_{t-2},\a_{t-1}\right)$. The augmented state will henceforth be referred to as \emph{information state}.

If an action $\a_t$ is selected at time $t$, the information state transitions to $\mathbf{I}_{t+1}=\left(\s_{t-d+1},\a_{t-d+1},\dots,\a_{t-1},\a_{t}\right)$ with the probability $p_{\a_t}\left(\s_{t-d+1}|\s_{t-d}\right)$. According to~\cite{altman1992closed}, the action $\a_t$ results in a reward $\mathbb{E}\left[r(\s_t,\a_t)|\mathbf{I}_t\right]$. Note that the reward structure depends on $\s_t$ and not $\s_{t-d}$, and thus evaluation of expected reward entails computation of the conditional probability distribution $p(\s_t|\mathbf{I}_t)$. The complexity of this computation grows as $O\left(d|\mathcal{S}|^2\right)$, which is why model-based methods~\cite{chen2020delay} are computationally prohibitive to work with if the dimension of state-space is large. Alternatively, one could try to reformulate CDMDP into another equivalent MDP with simplified reward structure such that while the two MDPs differ in their achievable optimal rewards, the underlying optimal policies are still the same~\cite{katsikopoulos2003markov}. Thus, one can simultaneously achieve both the accuracy of model-free methods, as well as sample-efficiency (characteristic of model-based methods). It must be remarked that despite the choice of reward-structure, augmentation is required to impart Markov property. The flavor of the simplified reward structure can be seen in the proof of Lemma~\ref{lem:equiv}, however, a detailed analysis is presented for the generalized setting of stochastic delay in Theorem~\ref{thm:RDMDP}.

\begin{lemma}[Equivalence of action and observation delays]\label{lem:equiv}
    The CDMDPs $\left\langle\mathcal{S},\mathcal{A},\mathcal{P}_\mathcal{A},r,\gamma,d,0\right\rangle$ and $\left\langle\mathcal{S},\mathcal{A},\mathcal{P}_\mathcal{A},r,\gamma,0,d\right\rangle$ are functionally equivalent to the MDP $\left\langle\mathcal{I},\mathcal{A},\mathcal{P}_\mathcal{A},r,\gamma\right\rangle$ with appropriately chosen information state $\mathcal{I}$.
\end{lemma}
\begin{proof}
    The proof is based on the fact that conditioned on the appropriately chosen information state, the optimal decision making for the two CDMDPs have a similar structure. Let us first consider the case of constant observation delay. We denote this observation delayed MDP by the CDMDP $\langle\mathcal{S},\mathcal{A},r,\gamma,d,0\rangle$. The information necessary for optimal action selection at any stage $t$ is given by $I_t=(\s_{t\!-\!d},\a_{t\!-\!d},\dots,\a_{t\!-\!2},\a_{t\!-\!1})$. According to~\cite{altman1992closed}, for an arbitrary but fixed policy $\pi:\mathcal{S}\times\mathcal{A}^d\to\mathcal{A}$, the total expected reward is given by:
\begin{align}\label{eq:AltmanV}
    V^\pi_\text{obs}(I_t) = \mathbb{E}_{\pi}\left[\sum\limits_{k\geq t}\gamma^{k-t}r(\s_k,\a_k)|I_t\right].
\end{align}
We consider another MDP with a modified cost-structure with the total expected reward given by:
\begin{align}\label{eq:ourV}
    \tilde{V}_\text{obs}^\pi(I_t) &= \mathbb{E}_{\pi}\left[\sum\limits_{k\geq t}\gamma^{k-t}r(\s_{k-d},\a_{k-d})|I_t\right] \nonumber \\
    &= \mathbb{E}_{\pi}\left[\sum\limits_{k\geq t-d}\gamma^{k-t+d}r(\s_{k},\a_{k})|I_t\right],
\end{align}
which can be further expanded as:
\begin{align}\label{eq:ourV1obs}
    \tilde{V}_\text{obs}^\pi(I_t) &= \mathbb{E}_{\pi}\left[r(\s_{t\!-\!d},\a_{t\!-\!d})+\gamma r(\s_{t\!-\!d\!+\!1},\a_{t\!-\!d\!+\!1})+\dots\right. \nonumber \\
    &\hspace{-2em}+ \left.\gamma^{d\!-\!1}r(\s_{t\!-\!1},\a_{t\!-\!1})|I_t\right] + \gamma^d\underbrace{\mathbb{E}_{\pi}\left[\sum\limits_{k\geq t}\gamma^{k-t}r(\s_k,\a_k)|I_t\right]}_{V^\pi_\text{obs}(I_t)}.
\end{align}
The first term on the right-hand side is independent of the policy $\pi$ when conditioned on the information state $I_t$. Therefore, it can be concluded that
\begin{align*}
    \argmax\limits_{\pi} \tilde{V}_\text{obs}^\pi(I_t) = \argmax\limits_{\pi} V_\text{obs}^\pi(I_t),
\end{align*}
i.e., the optimal policies for the two MDPs (with and without modified reward structure) are equivalent.

Next, we consider the CDMDP $\langle\mathcal{S},\mathcal{A},r,\gamma,0,d\rangle$ with action-delay. According to~\cite{altman1992closed}, this CDMDP can be reduced into an equivalent MDP with information state $I_t=(\s_{t},\a_{t\!-\!d},\dots,\a_{t\!-\!2},\a_{t\!-\!1})$ and the total expected reward given by:
\begin{align}\label{eq:AltmanVact}
    V^\pi_\text{act}(I_t) = \mathbb{E}_{\pi}\left[\sum\limits_{k\geq t}\gamma^{k-t}r(\s_{k\!+\!d},\a_{k})|I_t\right].
\end{align}
Following our analysis for the observation-delayed MDP with modified cost structure, we can obtain a similar relationship between the expected reward for the modified and unmodified rewards, i.e.,
\begin{align}\label{eq:ourV1act}
    \tilde{V}_\text{act}^\pi(I_t) &= \mathbb{E}_{\pi}\left[r(\s_{t},\a_{t\!-\!d})+\gamma r(\s_{t\!+\!1},\a_{t\!-\!d\!+\!1})+\dots\right. \nonumber \\
    &\hspace{-4em}+ \left.\gamma^{d\!-\!1}r(\s_{t\!+\!d\!-\!1},\a_{t\!-\!1})|I_t\right] + \gamma^d\underbrace{\mathbb{E}_{\pi}\left[\sum\limits_{k\geq t}\gamma^{k-t}r(\s_{k\!+\!d},\a_k)|I_t\right]}_{V^\pi_\text{act}(I_t)}.
\end{align}
As before, the policy that maximizes $\tilde{V}_\text{act}^\pi(I_t)$, also maximizes the ${V}_\text{act}^\pi(I_t)$. From the point of view of an agent taking an action, the agent just needs to work with the suitably modified reward structure and the corresponding information state, independent of the nature of delay. Finally, if we consider the CDMDP $\langle\mathcal{S},\mathcal{A},r,\gamma,d_o,d_a\rangle$, then for a decision maker, it suffices to consider an equivalent MDP with the information state $I_t=\left(\s_{t\!-\!d_o},\a_{t\!-\!d_o\!-\!d_a},\a_{t\!-\!d_o\!-\!d_a\!+\!1},\dots,\a_{t\!-\!1}\right)$.
\end{proof}
\begin{remark}
    Lemma~\ref{lem:equiv} suggests that it suffices to consider only one form of delay. Consequently, the rest of the paper takes into account scenarios only with delayed observation, since action delays can be similarly analyzed. Nonetheless, we later present some experiments with action delays for the sake of completeness.
\end{remark}

\subsection{Stochastic Delay Markov Decision Process}\label{subsec:SDMDP}
We now discuss the MDP with stochastic delays. Unlike in a CDMDP, in a stochastic delay MDP (SDMDP), the number of time steps between two successive observations is random and not equal to unity. Consequently, it is possible to observe a later state $\s_{t+1}$ before observing $\s_t$. This is in contrast to the SDMDP setting described in~\cite{katsikopoulos2003markov}, where it is assumed that $\s_{t+1}$ can only be observed after the state $\s_t$ has been observed.

For brevity, we only consider the scenario with random observation delays. The analysis for SDMDPs with action delays follows easily from Lemma~\ref{lem:equiv}. Below we formally define an SDMDP. 
\begin{definition}
    A Stochastic Delay Markov Decision Process (SDMDP), denoted by the tuple $\left\langle\mathcal{S},\mathcal{A},\mathcal{P}_\mathcal{A},r,\gamma,d_o,d_a,[k],n\right\rangle$, augments an MDP with state-space $\mathcal{S}\times\mathcal{A}^{d_o+d_a}\times\mathbb{Z}^+$ such that $d_o+d_a\leq n-1$. A policy $\pi$ in SDMDP is defined as a mapping $\pi:\mathcal{S}\times\mathcal{A}^{d_o\!+\!d_a}\times\mathbb{Z}^+\to\mathcal{A}\cup\{\emptyset\}$. Here $\emptyset$ represents \emph{no action} and corresponds to the scenario when the MDP \emph{freezes} for the agent. 
\end{definition}

Unlike with CDMDPs, in an SDMDP the information state $I_t$ must also include the time-instant $k$ at which the last observed system state $\s_{t\!-\!d_o}$ was first observed. This is required in order to ensure that the reward corresponding to $r(\s_{t\!-\!d_o},\a_{t\!-\!d_o})$ is collected only once at stage $k$. We also assume that if the delay for a later state $\s_{t+1}$ is smaller than the corresponding delay for $\s_{t}$ by more than unity, the agent gets to observe the most recent state $\s_{t+1}$ and the reward associated with it, along with the rewards associated with $\s_{t}$ and other previously unobserved states.

Another subtlety with SDMDPs is the size of information state $I_t$. Let $\s_{t\!-\!o}$ be the most recent observation first observed at instant $k$. Then for an agent trying to make an optimal decision, the information necessary at time $t$ is $I_t=\left(\s_{t\!-\!d_o},k,\a_{t\!-\!d_o},\a_{t\!-\!d_o\!+\!1},\dots,\a_{t\!-\!1}\right)$. If the state $\s_{t\!-\!d_o\!+\!1}$ becomes observable at the next instant, the system transitions to $I_{t\!+\!1}=\left(\s_{t\!-\!d_o\!+\!1},t+1,\a_{t\!-\!d_o+\!1},\a_{t\!-\!d_o\!+\!2},\dots,\a_{t\!}\right)$ with $|I_{t\!+\!1}|=|I_t|$. However, if the agent does not receive any new observation at time $t+1$, the system transitions to $I_{t\!+\!1}=\left(\s_{t\!-\!d_o},k,\a_{t\!-\!d_o},\a_{t\!-\!d_o\!+\!1},\dots,\a_{t}\right)$ with $|I_{t\!+\!1}|=|I_t|+1$. Thus, the size of the 
information state may vary during the evolution of an SDMDP. We account for this inconsistency by adding a `no action' to the action space and keeping the length of the information state at each stage to $n+1$ with the understanding that the maximum allowable delay is $n-1$. In scenarios where the delay amount exceeds $n-1$, it is assumed that the MDP freezes from the perspective of the agent, i.e., the agent does not take any new actions till the most recent state becomes observable. Thus at each instant, the information state $I_t$ is given by $(n+1)$~-~tuple $\left(\s_{t\!-\!d_o},k,\a_{t\!-\!d_o},\dots,\a_{t\!-\!1},\emptyset,\dots,\emptyset\right)$, where $\emptyset$ denotes the `no action'. In practice, `no action' can be regarded as equivalent of zero-padding when implementing deep-RL algorithms. Following the similar arguments as in \cite{altman1992closed, bertsekas1995dynamic}, it can be shown that SDMDP $\left\langle\mathcal{S},\mathcal{A},\mathcal{P}_\mathcal{A},r,\gamma,d_o,0,[k],n\right\rangle$ can be reduced to an undelayed MDP $\left\langle\mathcal{I},\mathcal{A},\mathcal{P}_\mathcal{A},r',\gamma\right\rangle$ with $r'(\s_t,\a_t)=\mathbb{E}_\pi\left[r(\s_t,\a_t)|I_t\right]$. The total expected reward for any arbitrary but fixed policy is thus given by
\begin{align}\label{eq:V-SDMDP}
    V^\pi_\text{obs}(I_t) = \mathbb{E}_\pi\left[\sum\limits_{m\geq t}\gamma^{m-t}r(\s_m,\a_m)\big|I_t\right].
\end{align}
As with CDMDPs, the reward structure in MDP in \eqref{eq:V-SDMDP} necessitates $O(d_o|\mathcal{S}^2|)$ computations which can be computationally prohibitive. The theorem below simplifies this cost structure by considering another MDP, whose optimal policy coincides with the optimal policy of the MDP in \eqref{eq:V-SDMDP}.
\begin{theorem}[Equivalence of SDMDP and undelayed MDP]\label{thm:RDMDP}
    The SDMDP $\left\langle\mathcal{S},\mathcal{A},\mathcal{P}_\mathcal{A},r,\gamma,d_o,0,[k],n\right\rangle$ can be reduced into an equivalent undelayed MDP $\left\langle\mathcal{I},\mathcal{A},\mathcal{P}_\mathcal{A},r,\gamma\right\rangle$ with simplified cost structure.
\end{theorem}
\begin{proof}
    Note that the set of instants $[k]$ at which the states are first observed do not explicitly enter into the accompanying technical arguments, however, it is needed to ensure that the reward collections occur only once per state.
    The proof of Theorem~\ref{thm:RDMDP} is similar to the derivation in Lemma~\ref{lem:equiv}. We consider an equivalent MDP where reward collection occurs as prescribed in Section~\ref{subsec:SDMDP}, i.e., if the delay at instant $t$ is $d_o$, then the corresponding reward $r(\s_t,\a_t)$ gets discounted by $\gamma^{t+d_o}$. Let $D$ be the observation delay random variable, i.e, $D$ takes values in the set $\{0,1,2,\dots,n-1\}$. Finally, let $\s_{t\!-\!d}$ be the last observable state that was first observed at stage $k\leq t$. The information state $I_t$ is given by $(\s_{t\!-\!d},k,\a_{t\!-\!d},\a_{t\!-\!d\!+\!1},\dots,\a_{t\!-\!1},\emptyset,\dots,\emptyset)$. The total expected reward for the MDP with modified reward structure is given by:
\allowdisplaybreaks
\begin{align}\label{eq:V-tilde-rd}
    \tilde{V}^\pi_\text{obs}(I_t) &= \mathbb{E}_{\pi,D}\bigg[\sum\limits_{m\geq t}\gamma^{m-t}r(\s_{m\!-\!D},\a_{m\!-\!D})\big|I_t\bigg] \nonumber \\
    &= \mathbb{E}_{\pi,D}\bigg[\sum\limits_{m\geq t-D}\gamma^{m-t+D}r(\s_{m},\a_{m})\big|I_t\bigg] \nonumber \\
    \tilde{V}^\pi_\text{obs}(I_t) &= \mathbb{E}_{\pi,D}\bigg[\sum\limits_{m\geq t-D}^{t-1}\gamma^{m-t+D}r(\s_{m},\a_{m})\big|I_t\bigg] \nonumber \\
    &\quad + \mathbb{E}_{\pi,D}\bigg[\sum\limits_{m\geq t}\gamma^{m-t+D}r(\s_{m},\a_{m})\big|I_t\bigg].
\end{align}
The second term on the right hand side of \eqref{eq:V-tilde-rd} conditioned on $I_t$ can be re-written as:
\begin{align*}
    \mathbb{E}_{D}\left[\gamma^D\right]\mathbb{E}_{\pi}\bigg[\sum\limits_{m\geq t}\gamma^{m-t}r(\s_{m},\a_{m})\big|I_t\bigg].
\end{align*}
Also, the first term on the right hand side of \eqref{eq:V-tilde-rd} is independent of $\pi$ conditioned on $I_t$. Also, since the policy fixed and independent of the delay process, it can be concluded that
\begin{align*}
    \argmax\limits_{\pi} V^\pi_\text{obs}(I_t) = \argmax\limits_{\pi} \tilde{V}^\pi_\text{obs}(I_t),
\end{align*}
which completes the proof.
\end{proof}

\section{Delay-Resolved Q-learning}\label{sec:DelayQ}
\begin{figure*}[h]
    \centering
    \subfloat[]{\includegraphics[scale=0.39]{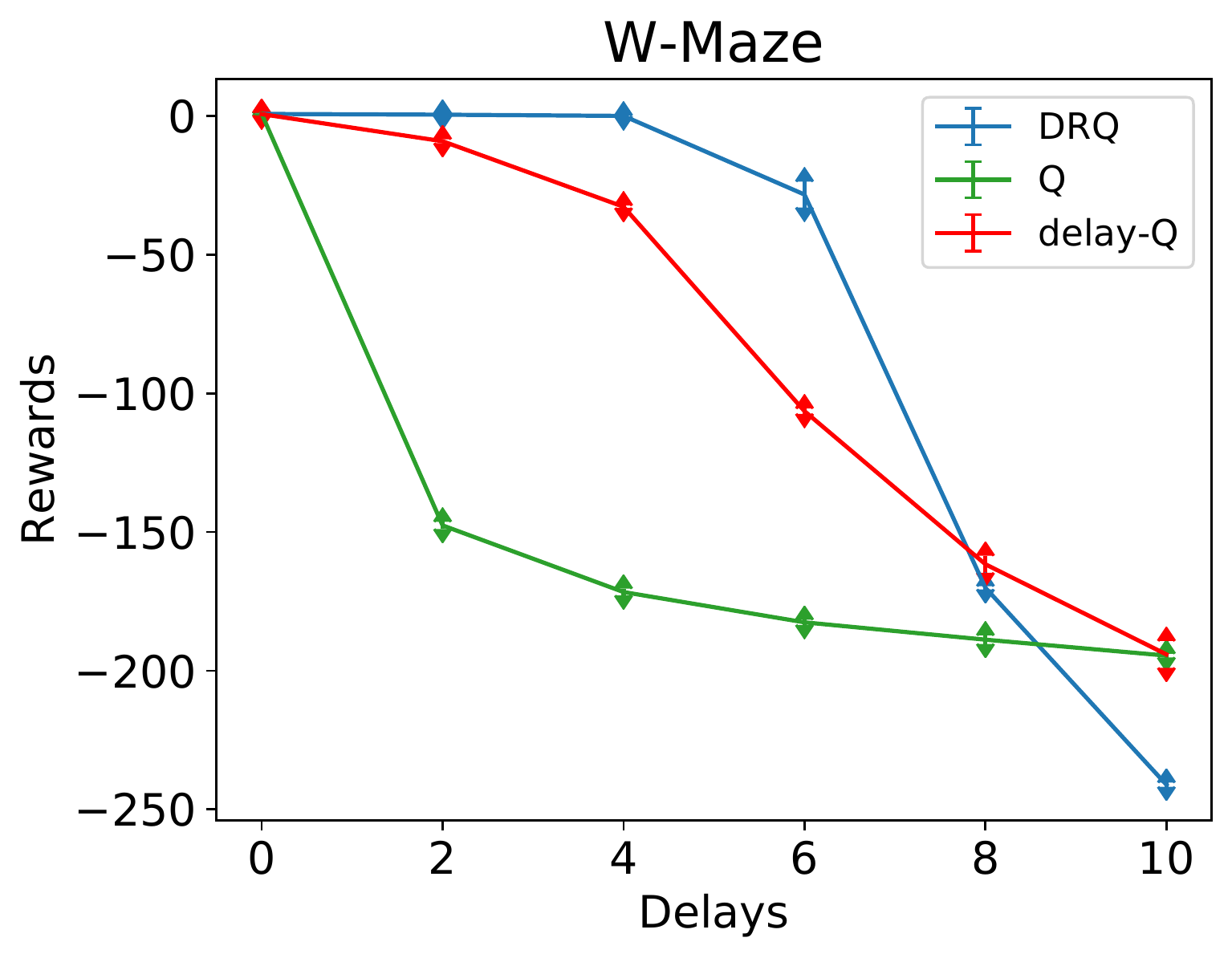}}
    \subfloat[]{\includegraphics[scale=0.39]{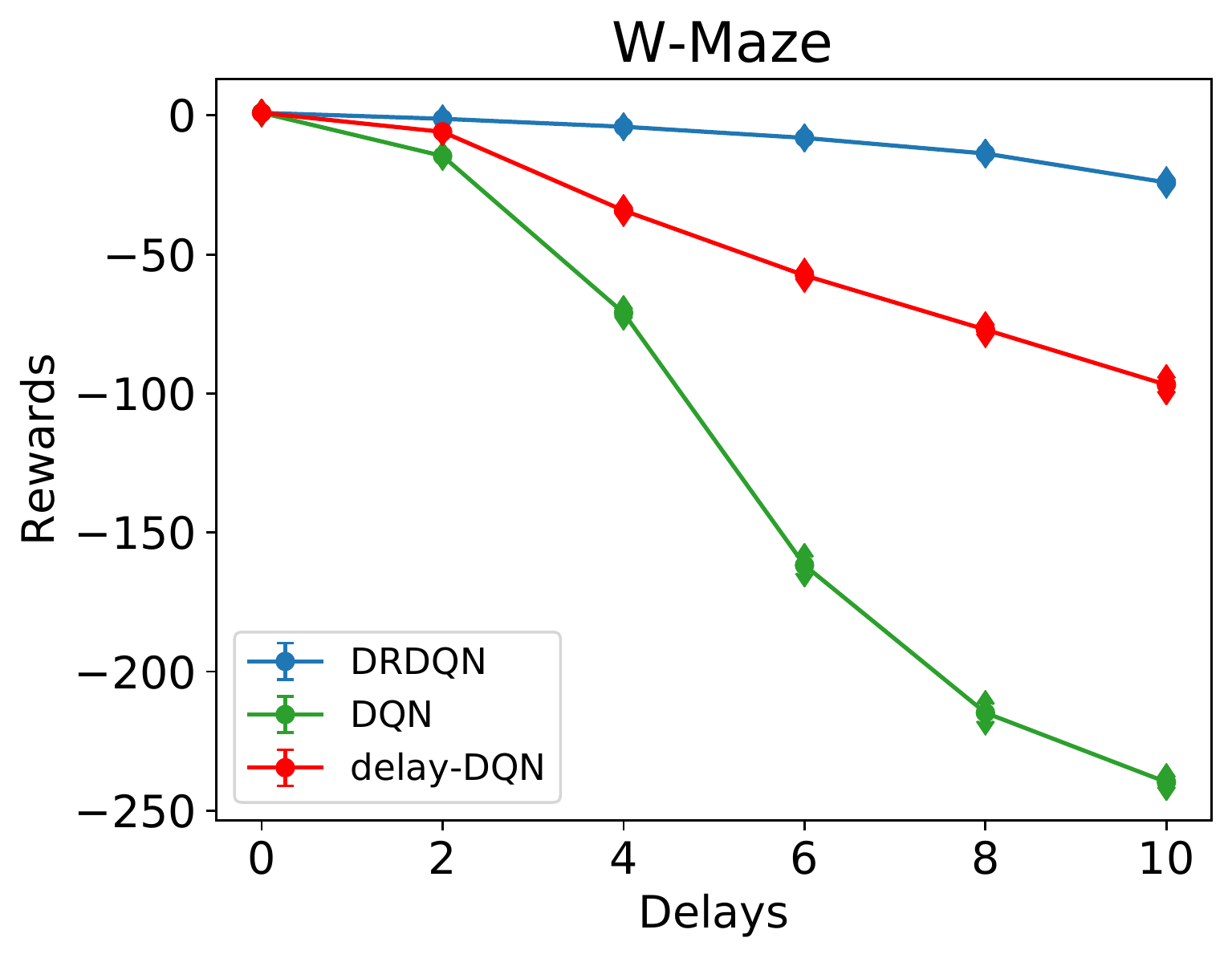}}
    \vspace{-1.0em}
    \caption{\centering Using Neural Networks to tackle the problem exploding state space: (a) shows the performance with Tabular Q-learning, whereas (b) uses DQN.}
    \vspace{-1.5em}
    \label{fig:w_maze_action}
\end{figure*}

The basic essence of the Delay Resolved algorithms is the conversion of non-Markov problems with both constant and stochastic delays (in actions and observations) into Markov problems, CDMDPs and SDMDPs respectively. Once MDPs have been formulated, we can apply any suitable reinforcement learning algorithm for both policy evaluation and improvement. 

The primary difference between Delay Resolved Q-learning and vanilla DQN (\cite{mnih2015human}) is the state-augmentation, i.e., instead of the state just being the current observation, there is an added history of pending actions which are stored in the \textit{action buffer} to make up an augmented state called \textit{information state}. For constant delays, the dimension of the information state, $|I_t|$ at time t, is constant. However, for stochastic delays, $|I_t|$ depends on the value of delay. As explained in Section \ref{subsec:SDMDP}, we assign a maximum allowable dimension for $|I_t|$  and fill up the \textit{action buffer} with `no-action'. Intuitively, it makes sense to choose the length of the \textit{action buffer} as the maximum possible delay-value if known a priori, however, compute requirements may impose additional restrictions on its length.

\subsection*{Comparison with Related Approaches}\label{sec:comp}
Though work on delayed RL environments is not very extensive, some approaches have been proposed to address the problem of decision making in presence of delays. Model-based RL approaches \cite{10.1007/978-3-540-74958-5_41,chen2020delay}, which involve learning the underlying undelayed transition probabilities can potentially be very useful if the dynamics are learned accurately. However, the learned models are often inaccurate, particularly in the presence of large or stochastic delays. Additionally, learning the state-transition probability is computationally expensive $(O\left(|\mathcal{S}|^2\right))$, which is precisely why model-based methods are not very popular in practice. Another recent approach \cite{Agarwal_Aggarwal_2021} used an expectation maximization algorithm where the agent takes decisions based on the expected next state, however it also requires estimating transition probabilities. Consequently, \cite{schuitema2010control} proposes a model-free approach that does not use augmented information states, instead, proposes to build an action history. Their scheme uses the effective action from the action history, typically the last action of the \textit{action buffer}, during the Q-learning update. Here, the effective action is the action that gets applied on the current observed state and makes the update more accurate than just incorporating the current pending action. However, their approach works only with action delays. Moreover, their framework is not delay-agnostic, i.e., \textit{the delay value must be known a priori.}

Most recently, \cite{dalal2021acting} addresses the CDMDP by using forward models for predicting the current undelayed state and selecting actions based on that predicted state. Though their algorithm shows promising results, it still is computationally more extensive than the normal model-free methods, because additional compute is required for learning the forward models. Learning good forward models is an essential aspect of this algorithm, which is generally not an easy task for all environments. Particularly with large delays, the forward model needs to be applied multiple times, which can lead to compounding errors. With stochastic delays, the task becomes even more difficult. Additionally, for best results, ideally the \textit{delay value will be necessary both during inference as well as training.} This method also requires additional storage in the form of storing a \textit{sample buffer} which is basically the trajectory of the pending actions along with their states and rewards. While, the delay resolved algorithm proposed in this paper requires additional memory for the \textit{action buffer}, it is much easier to store lower-dimensional actions than the potentially higher-dimensional trajectories.

\cite{ramstedt2020reinforcement} proposes a different solution for solving random delay problems by using augmented states along with the delay values for both action and observation. Their algorithm leverages the values of action and observation delays at each step to convert off-policy sampled trajectories into on-policy ones. Their algorithm, known as Delay Correcting Actor Critic (DCAC), performs reasonably well, however, at the expense of requiring \textit{complete information of the delay values at each step needed for re-sampling.} On the contrary, our DRDQN algorithm is designed to work in environments that do not leverage the additional information on delay values at each step. Consequently, DCAC is not suited for optimal decision-making in delay-agnostic environments discussed in this work.

\section{Experimental Results}\label{sec:exp}
\setcounter{figure}{6}
\begin{figure*}[h]
    \centering
    \subfloat[]{\includegraphics[scale=0.34]{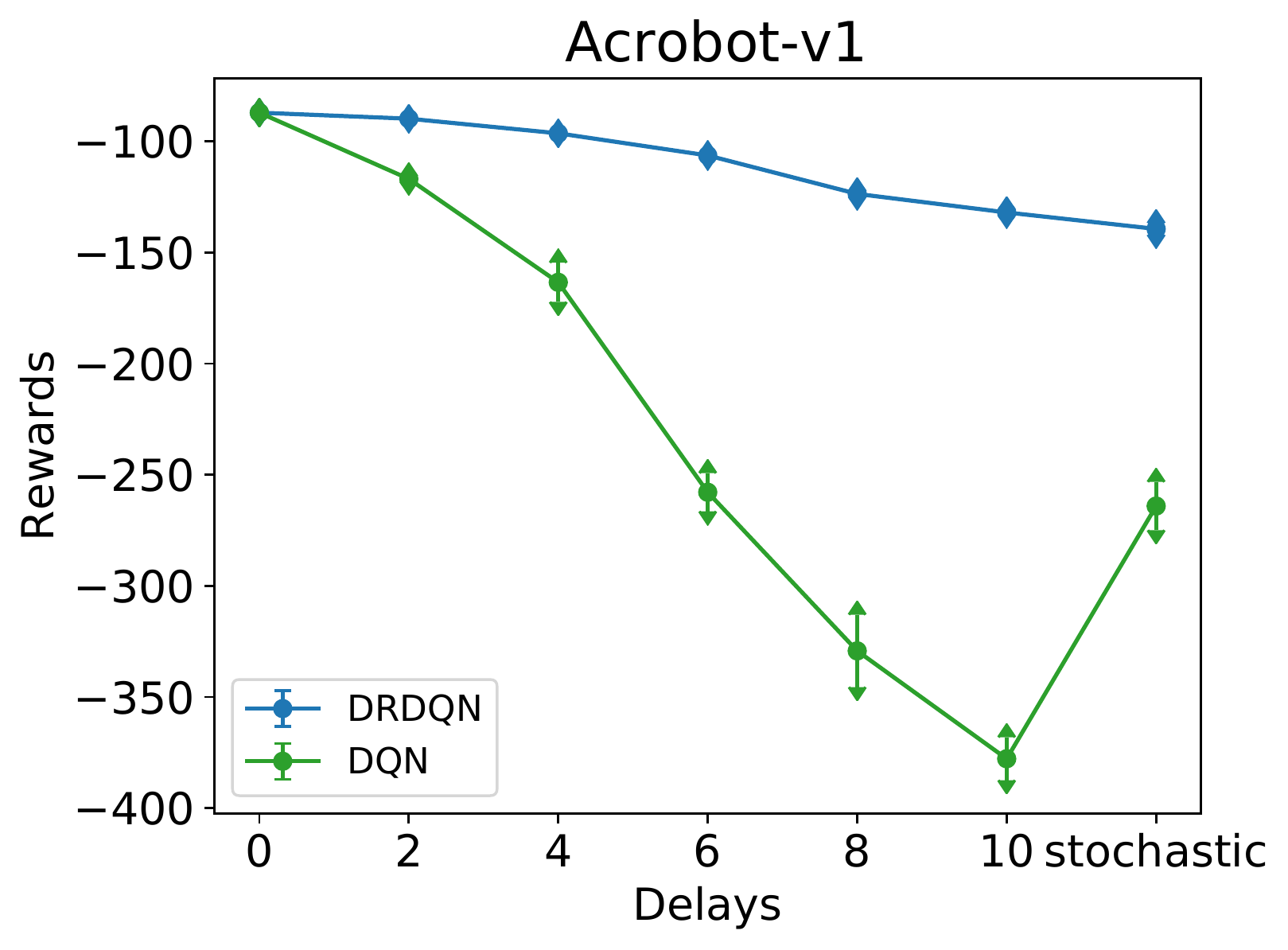}}
    \subfloat[]{\includegraphics[scale=0.34]{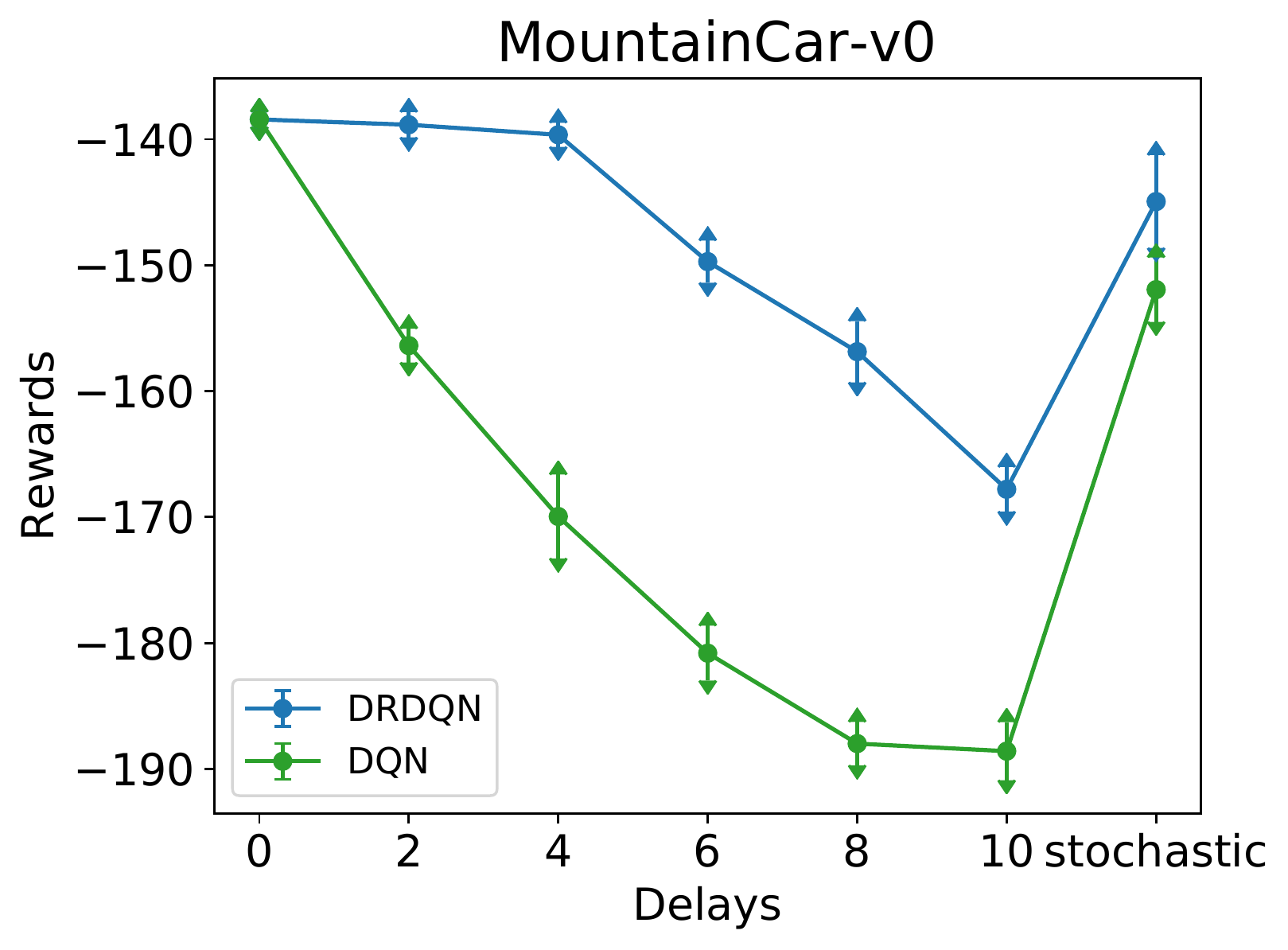}}
    \subfloat[]{\includegraphics[scale=0.34]{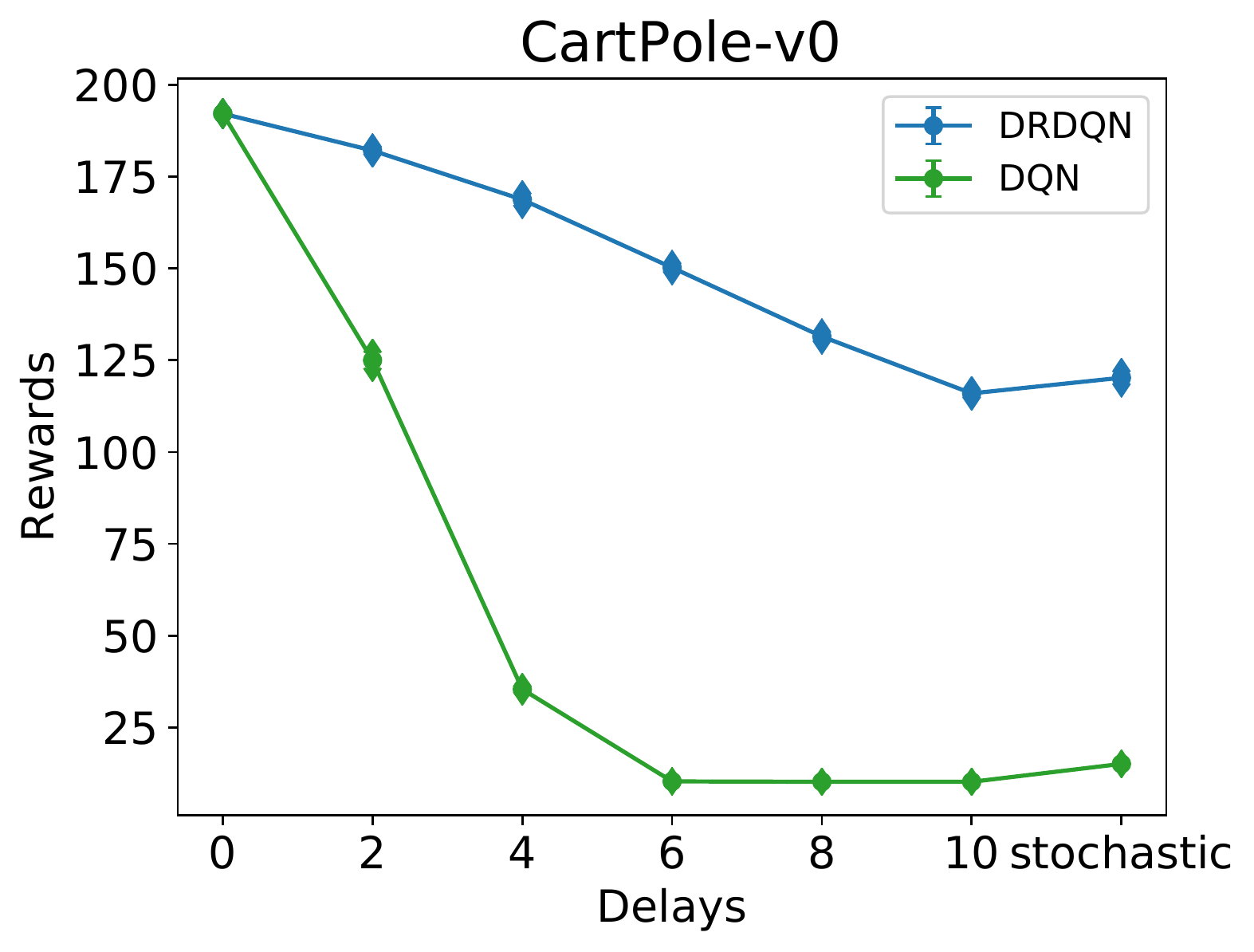}}
    \vspace{-0.5em}
    \caption{\centering Comparison with Baselines on Gym environments for constant and stochastic observation delays. DRDQN learns the optimal policy both for constant and stochastic delays.}
    \label{fig:obs_delay}
    \vspace{-1.5em}
\end{figure*}
\begin{figure*}[h]
    \centering
    \subfloat[]{\includegraphics[scale=0.34]{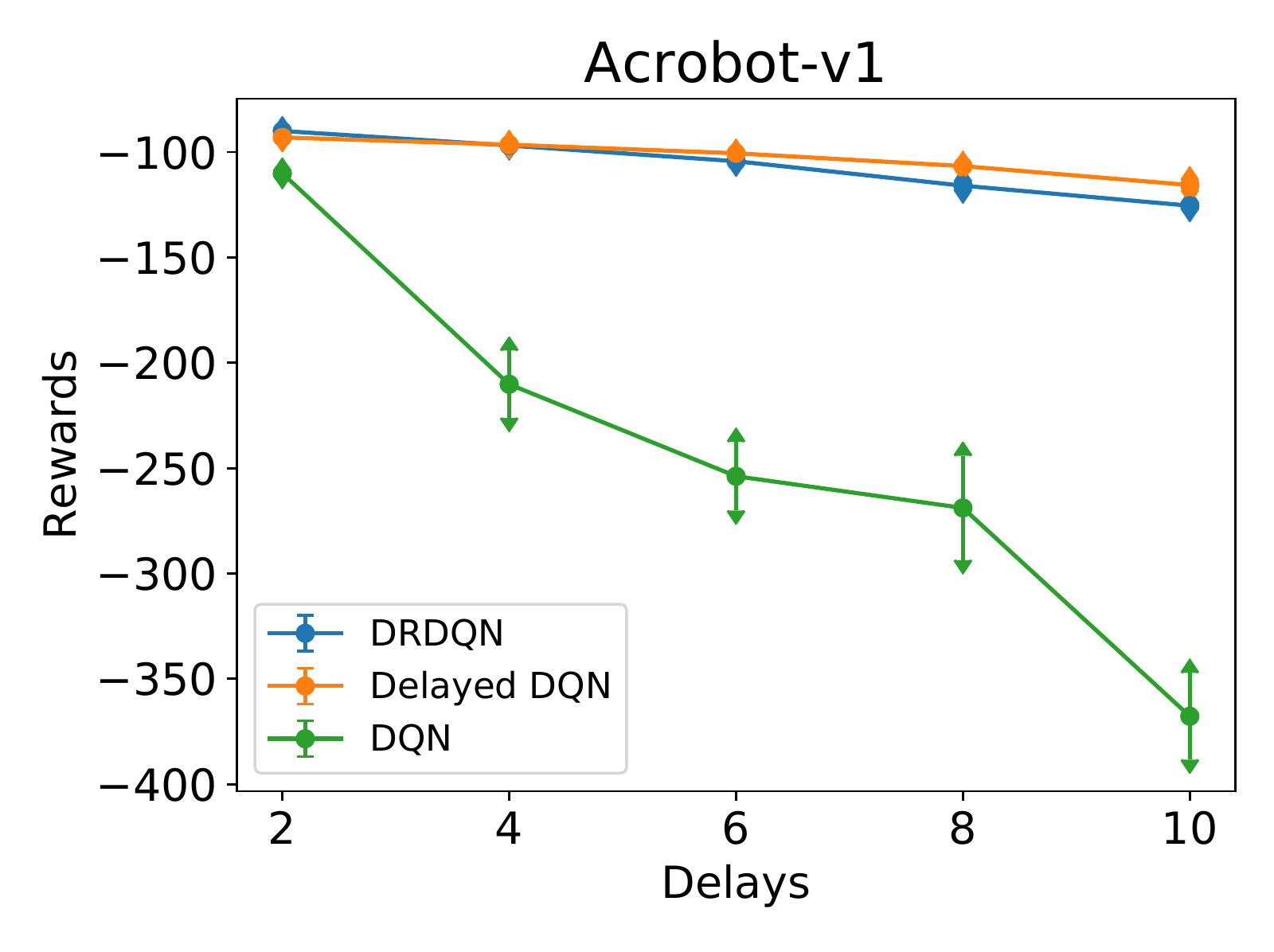}}
    \subfloat[]{\includegraphics[scale=0.34]{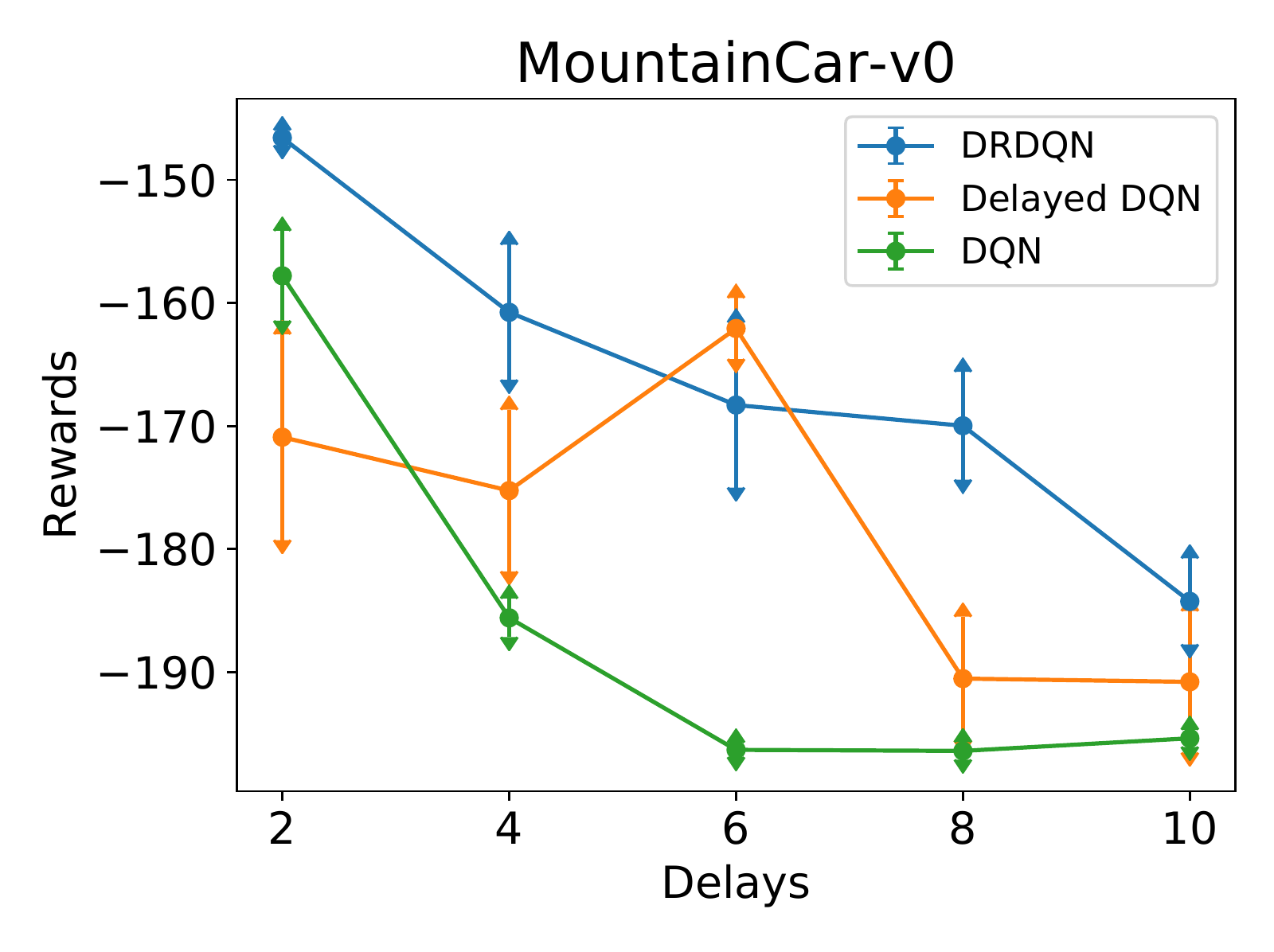}}
    \subfloat[]{\includegraphics[scale=0.34]{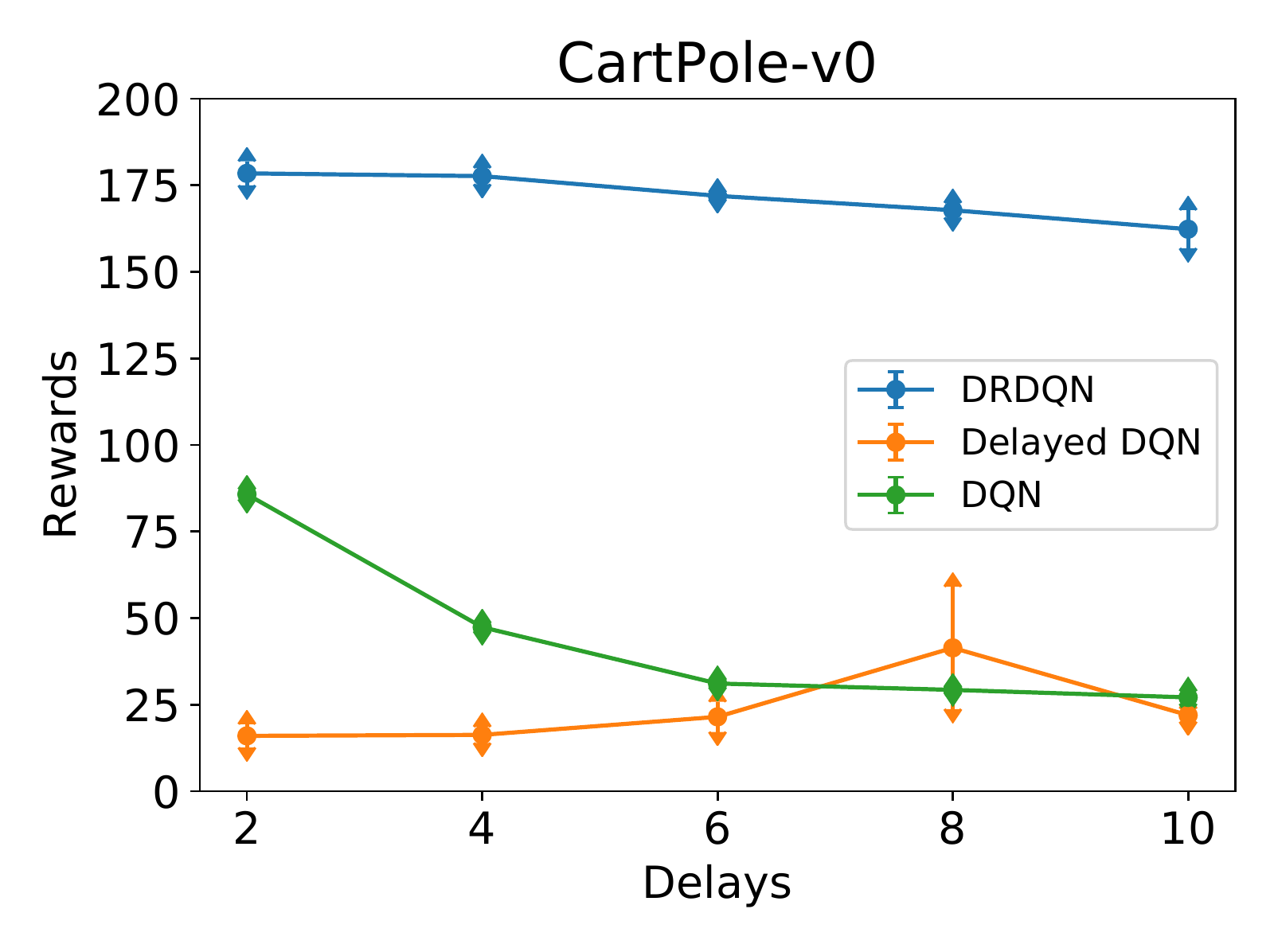}}
    \vspace{-0.5em}
    \caption{\centering Comparison with Baselines on Gym environments for constant action delays. In (a), both Delayed(\cite{dalal2021acting}) and DRDQN have similar performance, whereas in (b) and (c) due to inaccurate forward models, Delayed DQN does not perform as well.}
    \label{fig:action_delay}
    \vspace{-1.5em}
\end{figure*}

\begin{figure*}[h]
    \centering
     \subfloat[]{\includegraphics[scale=0.34]{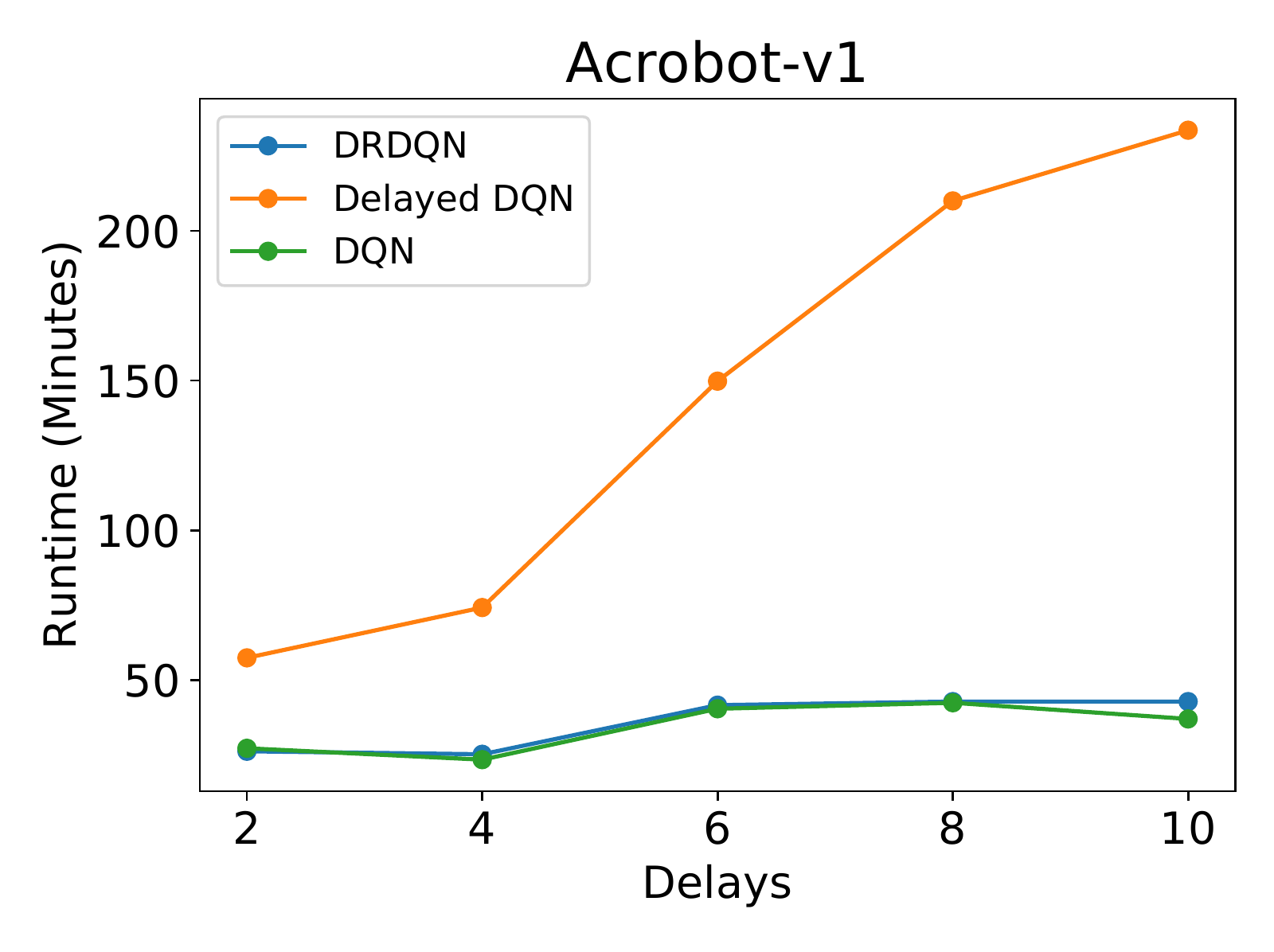}}
    \subfloat[]{\includegraphics[scale=0.34]{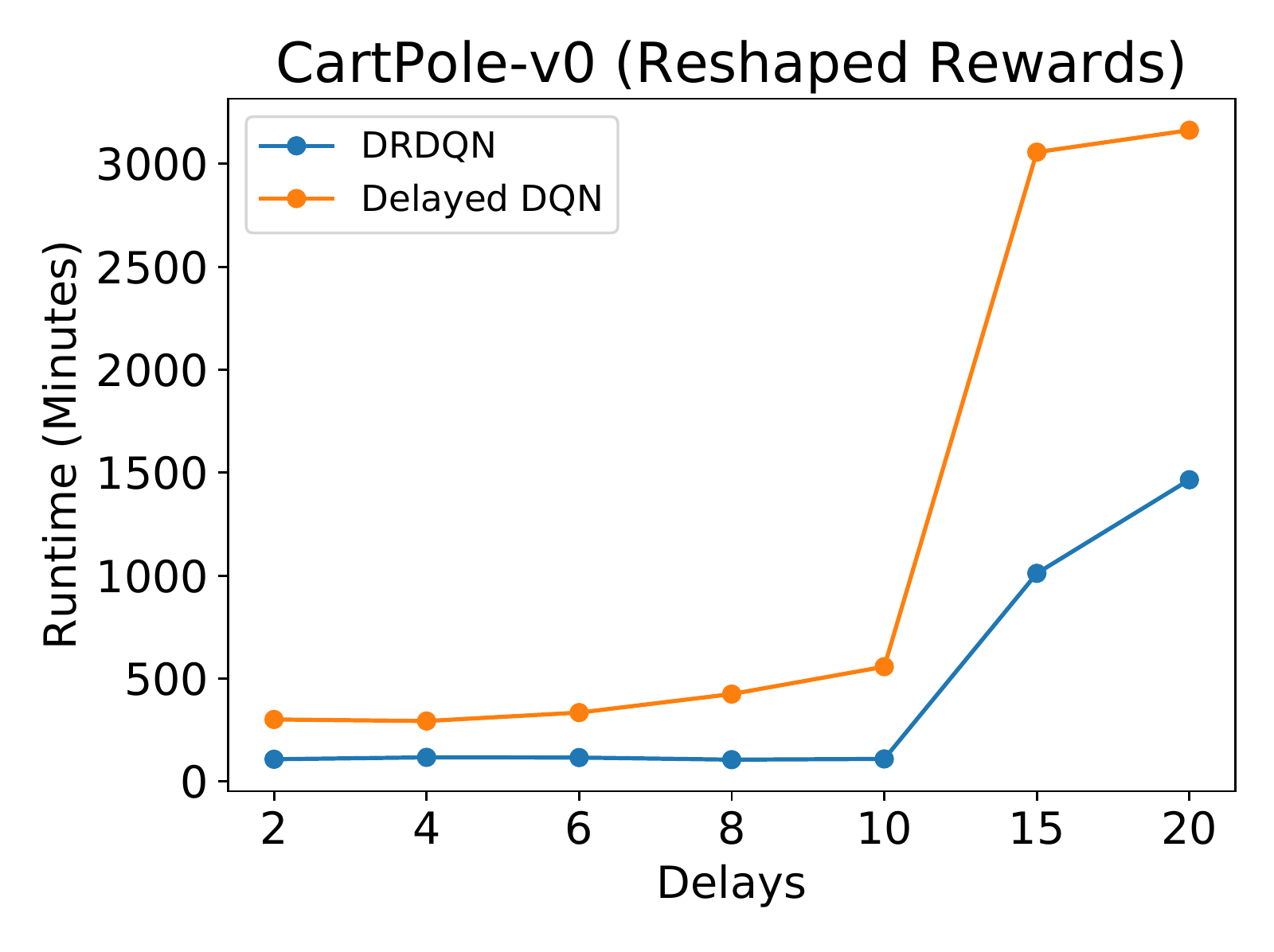}}
    \subfloat[ ]{\includegraphics[scale=0.34]{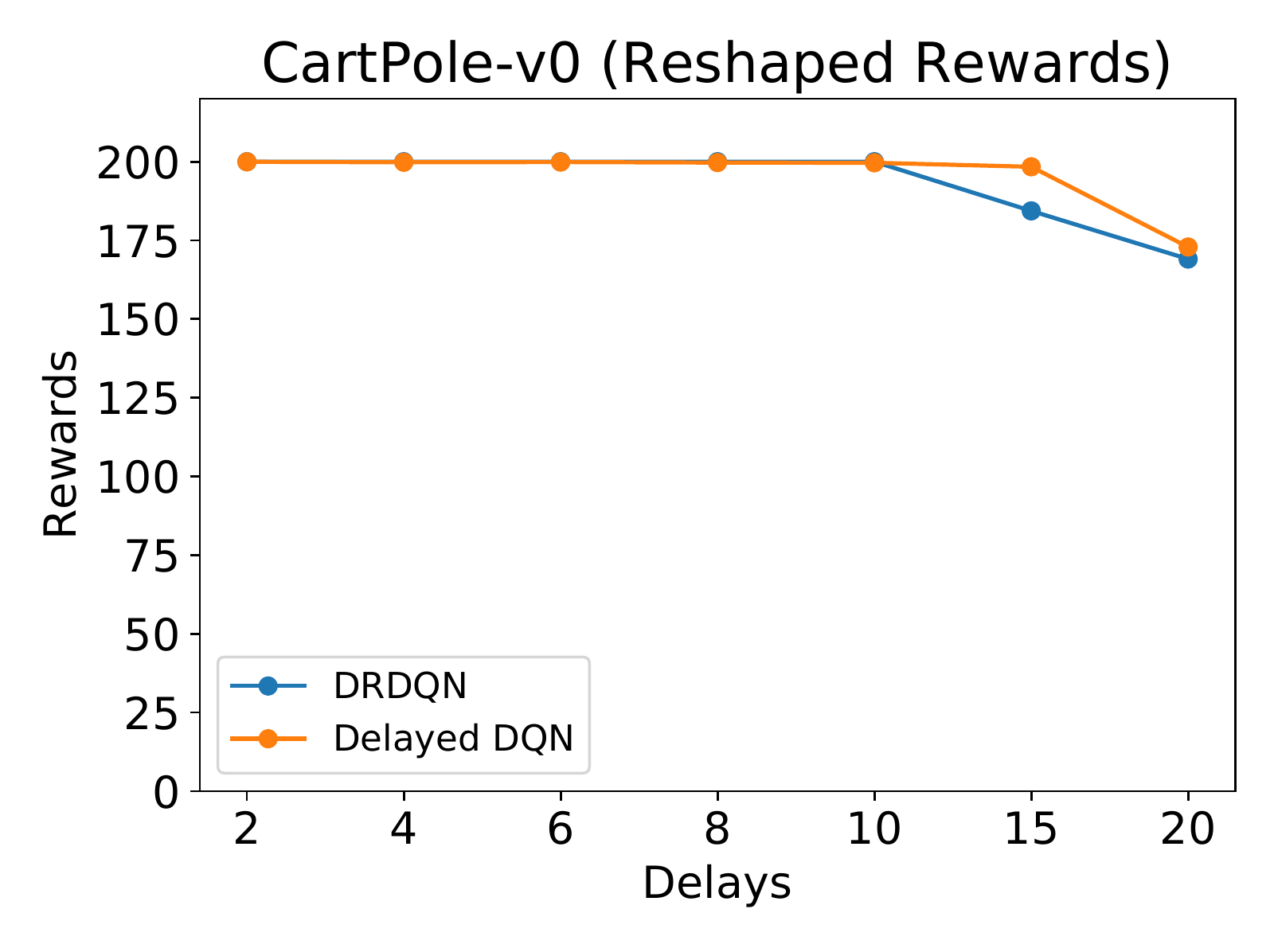}}
    \vspace{-0.5em}
    \caption{\centering (a) and (b) show runtimes for DRDQN and Delayed DQN(\cite{dalal2021acting}); (c) is asymptotic performance (average over last 10000 episodes)}
    \label{fig:time}
    \vspace{-1.5em}
\end{figure*}


We tested our algorithm on the delayed version of four standard environments, W-Maze, Acrobot, MountainCar and CartPole (details in \ref{sec:env}) with both action and observation delays. For all our experiments, we used Deep Q-Networks (\cite{mnih2015human}) as the base RL algorithm\footnote{Code available at: \url{https://github.com/baranwa2/DelayResolvedRL}}. Since, the primary modification is in how the state is structured and not on how the samples are collected and trained with, Delay Resolved algorithms can be used with any RL algorithm. 

\subsection{Environments}\label{sec:env}
We first describe the various environment dynamics along with the corresponding reward structures below.

\noindent \textbf{W-Maze}: W-Maze is a 7x11 grid world as shown in Fig. \ref{fig:w_maze_fig}. The agent starts randomly at the bottom row and the objective is to reach the GOAL states through available actions (UP, DOWN, LEFT, RIGHT). The agent receives a reward of +10 if it reaches the GOAL state and -1 for every other step not concluding at the GOAL state.
\setcounter{figure}{5}
\begin{figure}[H]
    \centering
    \vspace{-1em}
    \includegraphics[scale=0.3]{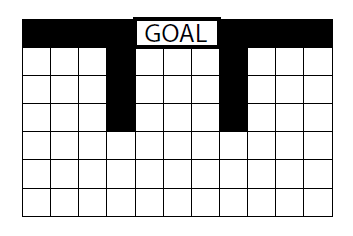}
    \vspace{-1.0em}
    \caption{W-Maze (\cite{schuitema2010control})}
    \vspace{-1.0em}
    \label{fig:w_maze_fig}
\end{figure}

\noindent \textbf{CartPole-v0}: In CartPole-v0, the agent aims to balance a pole on a cart by moving the cart either left or right. The agent receives a reward of +1 for every step and 0 when the episode ends, which happens when the agent either falls down or stays balanced for 200 time-steps. Thus the optimal reward for this environment is \textasciitilde{200}. Note that in the reshaped version of CartPole-v0, originally used in ~\cite{dalal2021acting}, the agent gets a reward based on the current value of its position and velocity and this reward grows as the pole becomes upright.

\noindent \textbf{Acrobot-v1}: Acrobot-v1 is a complex version of CartPole and comprises of two links and two independent actuators. In essence, Acrobot represents a double pendulum with an aim to swing the end of the lower-link up to a desired height. The actions comprise of torques generated by the actuator. The agent receives a step-reward of -1 and 0 once the episode ends. The optimal reward is \textasciitilde{-100}.

\noindent \textbf{MountainCar-v0}: In MountainCar-v0, the agent has to drive a vehicle uphill, however, the steepness of the slope is just enough to prevent it from driving up. Thus, he has to learn to drive backwards and build up momentum to reach the hilltop. The agent receives a step-reward of -1 and 0 on episode termination, which corresponds to him reaching the hilltop or 200 time-steps been elapsed.

\subsection{Exploding State Space and its Solution}

One of the biggest disadvantages of the algorithm as mentioned in ~\cite{dalal2021acting} is that the \textit{information state} space scales exponentially with delay value, $\mathcal{I}\coloneqq\mathcal{S}\times\mathcal{A}^d$. In tabular settings this becomes extremely problematic and infeasible to work with, specifically with large delays. However, if we use neural networks as function approximators, it is possible to include the \textit{action buffer} as an input concatenated with the state space, and thus the size of the input space becomes $|{\mathcal{S}}| + d|{\mathcal{A}}|$, which is easily manageable.

Fig~\ref{fig:w_maze_action} represents the average reward value of the last 10000 episodes along with their standard errors for 10 runs on the W-Maze environment~\cite{schuitema2010control} across various action delay values. From Fig.~\ref{fig:w_maze_action}a, it is evident that for smaller delays, delay resolved tabular Q-learning is able to achieve optimal reward (\textasciitilde{0}) for smaller delays. However, for large delays, they are not only computationally more expensive but also suffer from poor exploration. This is primarily due to exponential increase in the state-space resulting in poorer performance. The performance of the delay agent~\cite{schuitema2010control} is also depicted for the sake of completeness. This agent performs relatively well for smaller delays.

However, as shown in Fig.~\ref{fig:w_maze_action}b, neural networks as function approximator can handle the \textit{action buffer} as merely an input, which makes the \textit{information state} space much smaller and thus leads to optimal performance across larger delays as well. Interestingly, the delay-DQN algorithm proposed in~\cite{schuitema2010control}, performs reasonably well since we have a much powerful function approximation tool for predicting the value function. Although this is a useful trick, there are no theoretical guarantees on convergence to the optimal policy.

\subsection{Results with Constant and Stochastic Observation Delays}
A big advantage of Delay Resolved algorithms is that they can be used both for stochastic and constant delays, without requiring the delay input at each step as is needed for~\cite{ramstedt2020reinforcement}. Fig.~\ref{fig:obs_delay} highlights the performance of the proposed Delay Resolved DQN across three Gym environments. Each point is an average over the entire 1 Million steps of training across 10 runs (along with standard error bars), for constant as well as stochastic delays. The last column represents the stochastic delay uniformly selected between 0 and 10, with the maximum allowable dimension of the \textit{information state} being 10. Across all the environments, for both stochastic and constant delays, the DRDQN algorithm is able to converge to optimal policy as reflected in its near optimal rewards.

\subsection{Results with Constant Action Delays}
Fig.~\ref{fig:action_delay} portrays the average reward across the entire training regime (1 Million steps across 10 runs with the standard error bars), for various action delays in three different Open AI Gym~\cite{1606.01540} environments. For comparison, the average values for vanilla DQN~\cite{mnih2015human} and Delayed DQN~\cite{dalal2021acting} are exhibited alongside. As mentioned earlier, the delayed RL algorithm~\cite{dalal2021acting} is heavily influenced by the architecture of the forward model. For consistency, we used the same architecture as was used in the original paper~\cite{dalal2021acting} for the primary DQN agent. However, as seen in Figs.~\ref{fig:action_delay}b and \ref{fig:action_delay}c, it does not always work out well, and failing to learn an accurate forward model leads to worse performance. So, it is extremely critical to choose the correct forward model architecture and this will vary based on the environment, which can be very difficult to gauge ahead of training. Delay Resolved algorithms, on the other hand, do not involve learning any models and hence do not require any additional hyperparameter search. It does relatively well across all the domains for all the delay values.

Additionally, in Fig.~\ref{fig:action_delay}a, all the above algorithms perform relatively well getting to optimal rewards for large delays, however, the time taken is much longer for Delayed DQN~\cite{dalal2021acting} because of the additional training of the forward model. Fig.~\ref{fig:time}a depicts the time-taken for all the algorithms across all 10 runs. It can be clearly seen that Delay Resolved DQN is computationally very light and almost takes the same time as the vanilla DQN, whereas the Delayed DQN takes substantially longer per run for convergence.

\textbf{Note:} The points on each of the plots are an average over one million frames across 10 runs. Thus, the drop in performance for larger delays in Figs.~\ref{fig:obs_delay} and \ref{fig:action_delay} are due to the fact that the agent takes a longer time to reach the optimal performance which is reflected in the lower average score.

\subsection{Equivalence of Action and Observation Delays}\label{sec:equiv_act_obs}
Lemma~\ref{lem:equiv} establishes the equivalence of action and observation delays. For the sake of completeness, we plot the rewards obtained by DRDQN for the last 1000 episodes on Acrobot for both action and observation delays in Fig.~\ref{fig:equivalence}. It is evident that the DRDQN algorithm converges to similar values for action and observation delays across different values of delays.
\setcounter{figure}{9}
\begin{figure}[!ht]
    \centering
    \vspace{-1em}
    \includegraphics[scale=0.3]{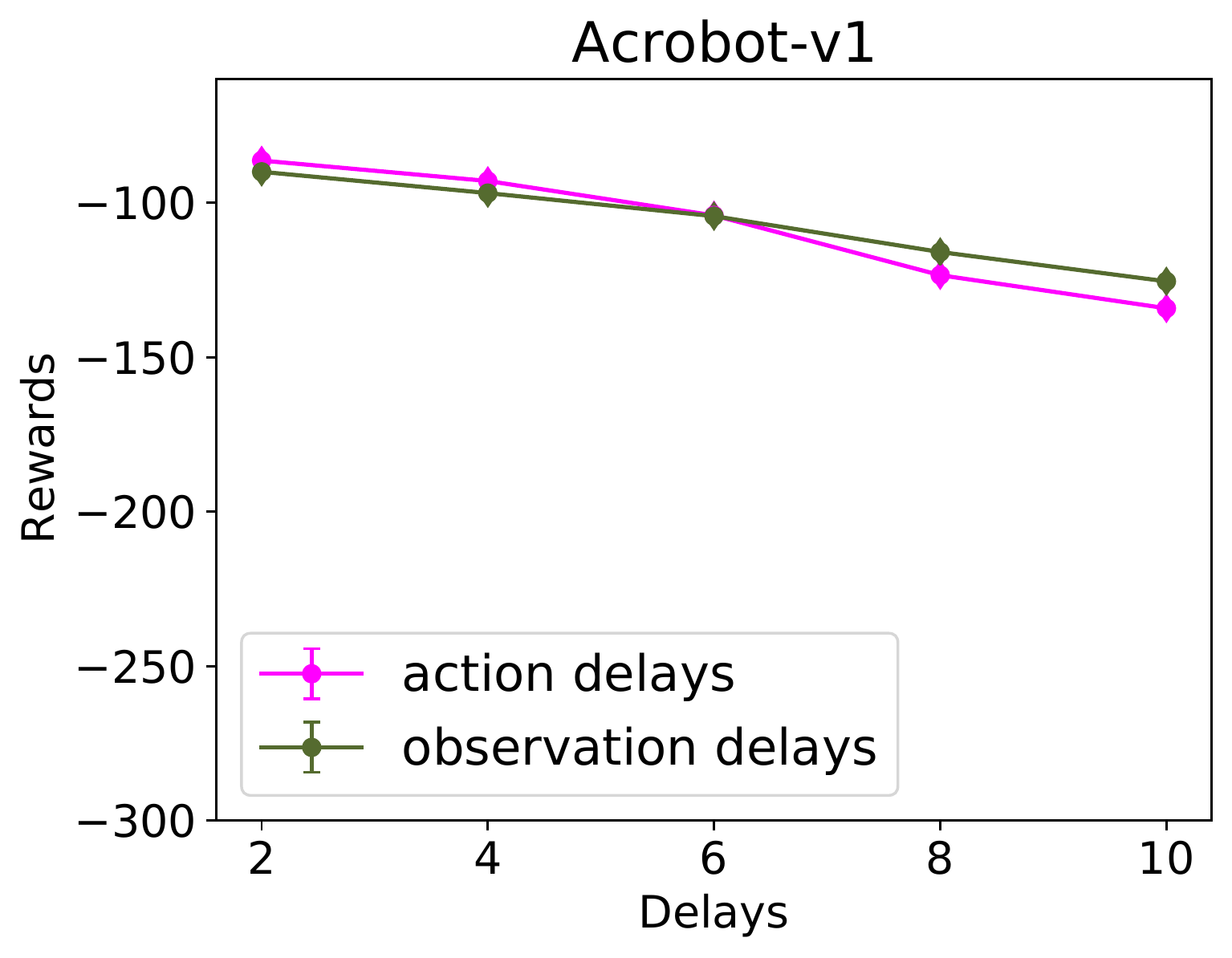}
    \vspace{-1.5em}
    \caption{Equivalence of Action and Observation Delays}
    \label{fig:equivalence}
    \vspace{-1em}
\end{figure}
\subsection{Compute Comparison}{\label{sec:compute}}
The only modification of Delay Resolved algorithms is in the addition of \textit{action buffer} in the state space. Although, this addition can lead to exponential increase of compute in tabular settings, when added as an input to a neural network value function approximator, the additional computational overhead is very minimal as can be seen in Figs.~\ref{fig:time}a and b. With the increase in delay, the run-time increases dramatically for Delayed method ~\cite{dalal2021acting} as the agent has to first compute the forward model and then apply the forward model multiple times in order to predict the current state. The environment used for Figs.~\ref{fig:time}b and c is the same as CartPole-v0 except with reshaped rewards as used in~\cite{dalal2021acting} (details in Section~\ref{sec:env}). We see that if appropriate forward models also reach optimality if appropriate architecture is chosen, but the time required is larger (Fig.~\ref{fig:time}b).


\section{Conclusion and Future Work}\label{sec:conclusion}

In this work, we have addressed a very crucial assumption in Reinforcement Learning, which is the synchronization between the environment and the agent. In real world scenarios, we cannot expect this assumption to hold true and thus we would need to modify our RL algorithms accordingly. This paper revisits the idea of using augmented states as a solution for delayed RL problems. We formally define both constant and stochastic delays and provide theoretical analysis on why the reformulation still converges to the same optimal policy. Additionally, we adapt state augmentation methods to constant and stochastic delay problems, to formulate a class of Delay Resolved RL algorithms, which can perform well on both constant and random delay tasks. 

For future work, we would like to extend this algorithm to real-world problems including accounting for actuator and motor delays in Robotics, and shipping delay in Supply Chains. Delay Resolved algorithms could prove to be a founding block for practical real-time RL algorithms. For example, without relaxing the assumption of knowing the delay value at every time-step, Delay Resolved DQN can still be improved upon by using better sampling strategies from the experience replay buffer.

\clearpage
\bibliographystyle{ACM-Reference-Format}
\bibliography{bibliography}

\end{document}